\documentclass[lettersize,journal]{IEEEtran}
\usepackage{amsmath,amsfonts}
\usepackage{algorithmicx}
\usepackage{algorithm}
\usepackage{array}
\usepackage{textcomp}
\usepackage{stfloats}
\usepackage{url}
\usepackage{verbatim}
\usepackage{graphicx}
\usepackage{cite}
\usepackage{arydshln}

\usepackage{algpseudocode} 
\usepackage{multirow}%
\usepackage{amsmath,amssymb,amsfonts}%
\usepackage{amsthm}%
\usepackage{mathrsfs}%
\usepackage{manyfoot}%
\usepackage{threeparttable}
\usepackage{booktabs}
\usepackage{subcaption}
\usepackage{hyperref}

\begin{document}

\title{ESVAE: An Efficient Spiking Variational Autoencoder with Reparameterizable Poisson Spiking Sampling}

\author{Qiugang~Zhan, 
        Ran~Tao,
        Xiurui~Xie, 
        Guisong~Liu,~\IEEEmembership{Member,~IEEE},
        Malu~Zhang,~\IEEEmembership{Member,~IEEE},
        Huajin~Tang,~\IEEEmembership{Senior Member,~IEEE},
        and~Yang~Yang,~\IEEEmembership{Senior Member,~IEEE}

        \thanks{Manuscript received XX, X; revised XX, X.}
        
        \thanks{This work was supported by the National Natural Science Foundation of China (NSFC) (NO. 62376228), and Chengdu Science and Technology Program (NO. 2023-JB00-00016-GX). 
        \textit{(Corresponding author: Guisong Liu and Xiurui Xie.)}}

        \thanks{Qiugang Zhan, Ran Tao, and Guisong Liu are with the Complex Laboratory of New Finance and Economics, School of Computing and Artificial Intelligence, Southwestern University of Finance and Economics, Chengdu, 611130, China (email: gliu@swufe.edu.cn).}

        \thanks{Xiurui Xie is with the Laboratory of Intelligent Collaborative Computing, University of Electronic Science and Technology of China, Chengdu, 611731, China (email: xiexiurui@uestc.edu.cn).}

        \thanks{Malu Zhang and Yang Yang are with the School of Computer Science and Engineering, University of Electronic Science and Technology of China, Chengdu, 611731, China.}

        \thanks{Huajin Tang is with the College of Computer Science and the State Key Laboratory of Brain-Machine Intelligence, Zhejiang University, Hangzhou 310027, China.}
        }

\markboth{Journal of \LaTeX\ Class Files,~Vol.~14, No.~8, August~2021}%
{Shell \MakeLowercase{\textit{et al.}}: A Sample Article Using IEEEtran.cls for IEEE Journals}

\IEEEpubid{0000--0000/00\$00.00~\copyright~2021 IEEE}

\maketitle

\begin{abstract}
  In recent years, studies on image generation models of spiking neural networks (SNNs) have gained the attention of many researchers. 
  Variational autoencoders (VAEs), as one of the most popular image generation models, have attracted a lot of work exploring their SNN implementation. 
  Due to the constrained binary representation in SNNs, existing SNN VAE methods implicitly construct the latent space by an elaborated autoregressive network and use the network outputs as the sampling variables. 
  However, this unspecified implicit representation of the latent space will increase the difficulty of generating high-quality images and introduce additional network parameters. 
  In this paper, we propose an efficient spiking variational autoencoder (ESVAE) that constructs an interpretable latent space distribution and designs a reparameterizable spiking sampling method. 
  Specifically, we construct the prior and posterior of the latent space as a Poisson distribution using the firing rate of the spiking neurons. 
  Subsequently, we propose a reparameterizable Poisson spiking sampling method, which is free from the additional network. 
  Comprehensive experiments have been conducted, and the experimental results show that the proposed ESVAE outperforms previous SNN VAE methods in reconstructed \& generated image quality. 
  In addition, experiments demonstrate that the encoder of ESVAE can retain the original image information more efficiently and is more robust. 
  The source code is available at \href{https://github.com/QgZhan/ESVAE}{https://github.com/QgZhan/ESVAE}.
\end{abstract}

\begin{IEEEkeywords}
Spiking neural network, Variational autoencoder, Image generation, Reparameterization trick.
\end{IEEEkeywords}

\section{Introduction}
\IEEEPARstart{R}{ecently}, artificial intelligence-generated content (AIGC) has become a popular research topic in both academic and business communities \cite{zhang2023complete}. 
Variational autoencoder (VAE) is one of the most popular image generation models and has been proven to be powerful on traditional artificial neural networks (ANNs) \cite{kingma2013auto, liu2021domain, shamsolmoali2023vtae}.  
However, it comes with a substantial computational power consumption, which makes it extremely challenging to implement AIGC on low-resource edge devices \cite{xu2023unleashing}. 
Therefore, researchers have started to explore the implementation of VAE on spiking neural networks (SNNs).

As the third generation neural network, SNNs achieve extremely low computational power consumption by simulating the structure of biological brain neurons \cite{maass1997networks, zhan2022bio, liu2022human, chakraborty2021fully}.   
Information propagation in SNN is through the spiking signals emitted by neurons, represented by binary time series data \cite{zhan2024spiking, yao2023sparser}. 
Relying on this hardware-friendly communication mechanism, SNN models are easy to be implemented by neuromorphic chips such as Loihi \cite{davies2018loihi}, TrueNorth \cite{debole2019truenorth} and Tianjic \cite{pei2019towards}. 

The critical ideas of VAEs are to construct the latent space distribution and sample latent variables to generate images, which are also the main challenges of SNN VAEs. 
In traditional ANN VAEs, the posterior distribution of the latent space is inferred by an encoder, and the prior distribution is usually preset to a Gaussian distribution \cite{kingma2013auto}. 
The reparameterization trick is introduced to make the latent variables sampling differentiable. 
However, for SNN, it is difficult to predict the parameters of Gaussian distribution with binary sequences, as well as to sample spiking latent variables with existing reparameterization tricks. 

To address these issues, some works propose hybrid SNN-ANN autoencoders that consist of an SNN encoder and an ANN decoder \cite{skatchkovsky2021learning, stewart2022encoding}. 
\cite{talafha2020biologically} proposed a pure SNN VAE model SVAE by converting a trained ANN model into the SNN version. 
FSVAE is the first VAE model constructed by fully SNN layers and can be directly trained without ANNs \cite{kamata2022fully}. 
FSVAE uses an autoregressive SNN model to construct the latent space, and sample variables from the model outputs using Bernoulli processes. 
Although TAID introduces an attention mechanism into FSVAE to improve the image quality, it does not propose new hidden space construction and sampling methods \cite{qiu2023when}.
In general, existing SNN VAE models either rely on ANNs for training or construct the latent space implicitly through additional network structures.

\IEEEpubidadjcol 

In this paper, we propose an efficient spiking variational autoencoder (ESVAE) in which the latent space is explicitly constructed by Poisson distributions. 
The Poisson-based prior and posterior distribution of latent space is represented by the firing rate of neurons. 
Then, we further propose the reparameterizable Poisson spiking sampling method to achieve a broader random sampling range than FSVAE. 
To avoid the non-differentiable problem arising from the sampling process, we introduce the surrogate gradient strategy of SNN training so that the proposed ESVAE model can be trained based on back-propagation. 
The experimental results demonstrate that the quality of both image reconstruction and generation of ESVAE exceeds that of extant SNN VAE methods.

The main contributions of this work are summarized as follows:
\begin{itemize}
    \item  We propose an ESVAE model, which explicitly constructs the latent space in SNN based on the Poisson distribution, improving the quality of the generated image and enhancing the interpretability of SNN VAE. 
    \item A Poisson spike sampling method is proposed, which is non-parametric and has a broader sampling range. It comes with a reparameterization method that is well-compatible with SNN training. 
    \item The ESVAE model is experimentally verified to have higher image reduction ability, stronger robustness, and better encoding ability than the previous SNN VAE model. 
\end{itemize}

\section{Background}
In this section, we briefly introduce the spiking neuron model and explore the temporal robustness of spiking latent variables. 

\subsection{Spiking Neuron Model}
The Leaky integrate-and-fire (LIF) model is one of the most widely used SNN neuron models \cite{wu2019direct, xie2024federated}. 
The LIF model integrates the membrane potential over time as influenced by input spiking sequences, and emits a spike when the membrane potential surpasses the threshold $v_{\theta}$. 
The entire process comprises three phases: charging, firing, and resetting, governed by:
\begin{equation}
    m^{i, t} = \frac{1}{\tau}v^{i, t-1} + \sum_{j} w^{i,j} o^{j, t},
\end{equation}
\begin{equation}
    o^{i, t} = H \left( m^{i, t}; v_{\theta} \right) = \begin{cases}
        1, & m^{i, t} \geq v_{\theta}, \\ 
        0, & m^{i, t} < v_{\theta}, 
    \end{cases}
    \label{eq: firing function}
\end{equation}
\begin{equation}
    v^{i, t} = m^{i, t} \left( 1 - o^{i, t} \right) + v_{\text {reset}} o^{i, t},
\end{equation}
where $o^{j, t}$ denotes the spike generated by neuron $j$ in previous layer at the $t^{th}$ time step. 
Neuron $i$ integrates the weighted spiking input from the previous layer with its membrane potential $v^{i, t-1}$ at the $t^{th}$ time step to derive the current instantaneous membrane potential $m^{i, t}$, where $w^{i,j}$ denotes the synaptic weight between neurons $i$ and $j$. 
$\tau$ represents the membrane potential decay factor. 
$H(\cdot)$ is a Heaviside step function that determines whether the output $o^{i, t}$ of neuron $i$ at the $t^{th}$ time step is 0 or 1. 
Based on $o^{i, t}$, the membrane potential $v^{i, t}$ of neuron $i$ is set to either the instantaneous membrane potential $m^{i, t}$ or the reset potential $v_{\text{reset}}$. 

\begin{figure}[htbp]
    \centering
	\includegraphics[scale=0.6]{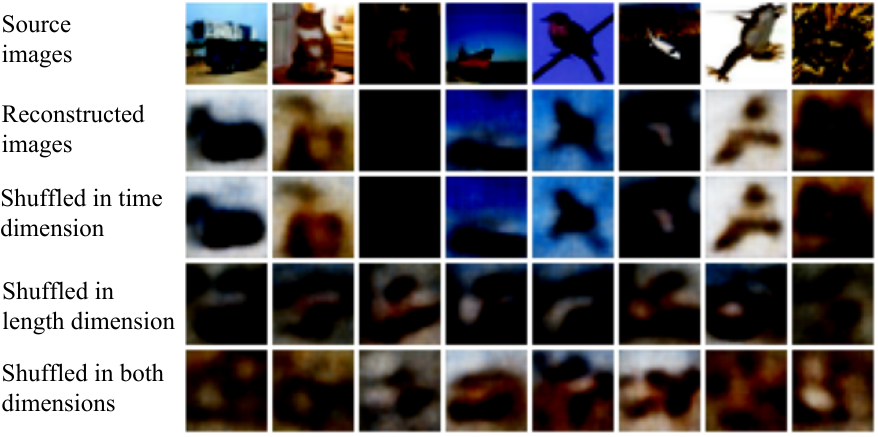}
    \caption{Comparison of vanilla reconstructed images and images generated by different latent variables on CIFAR10. }
    \label{fig: out-of-order}
\end{figure}

\subsection{Temporal Robustness in Spiking Latent Variables} \label{sec: temporal robustness}
To rationally construct the spiking latent space explicitly, in this section, we analyze how the latent variables affect the generated images on FSVAE \cite{kamata2022fully}.

In a general VAE, the posterior distribution $p\left(z|x\right)$ of the latent space is constructed for each input $x$. 
A latent variable $z$ is randomly sampled from $p\left(z|x\right)$ and fed into the generative distribution $p\left(x|z\right)$ implemented by a decoder network. 
For SNN VAE, the latent variables $z \in \{0, 1\}^{d, T}$ are a set of binary spike sequences, where $d$ is the length and $T$ is the SNN time window. 

To investigate the effect of latent spiking variables, we sample a latent spiking variable $z $ using the autoregressive Bernoulli method of FSVAE. 
We then shuffle the spikes along the length and time dimensions, respectively. 
Fig. \ref{fig: out-of-order} shows the comparison of different generated images on CIFAR10. 
Through this experiment, we discover the temporal robustness phenomenon: shuffling in the time dimension has negligible effect, while shuffling in the length dimension significantly changes the generated images. 

Further reflection on this phenomenon reveals that disrupting spikes in the time dimension does not change the firing rate of each latent variable neuron. 
This discovery lays the foundation for constructing latent spaces with Poisson distributions, which will be detailed in Sec. \ref{sec: distribution}. 

\begin{figure*}[htbp]
    \centering
	\includegraphics[scale=0.43]{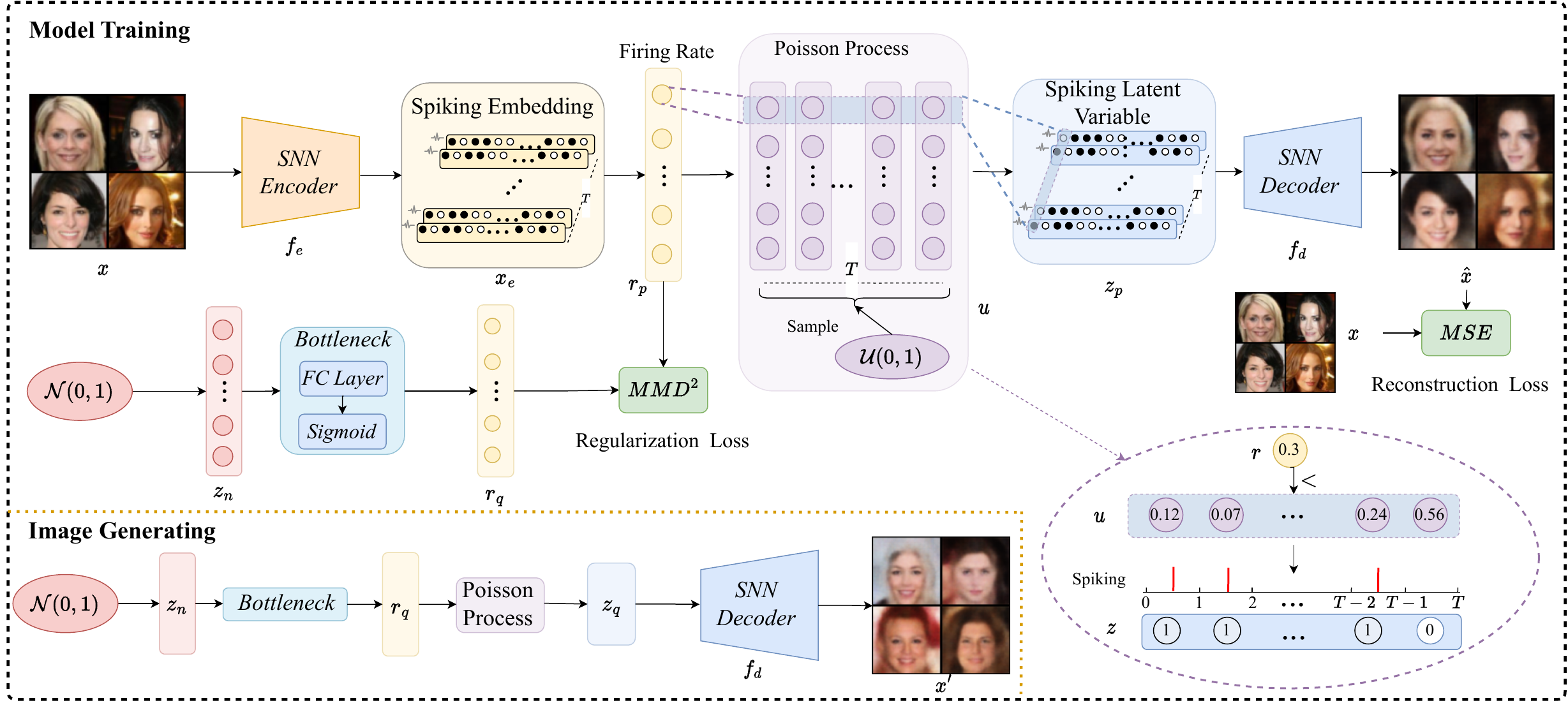}
    \caption{The model training and image generating processes of ESVAE. }
    \label{fig: ESVAE}
\end{figure*}

\section{Methods}
In this section, we propose an efficient spiking variational autoencoder (ESVAE) that uses a more straightforward reparameterizable Poisson sampling method. 
We introduce the construction of the posterior and prior distributions of the latent space in Sec. \ref{sec: distribution}. 
Then, the proposed Poisson spiking sampling method is described in Sec. \ref{sec: Poisson spiking sampling}. 
The evidence lower bound (ELBO) and loss function are derived in Sec. \ref{sec: ELBO and loss}. 

The training and image generating processes are shown in Fig. \ref{fig: ESVAE}.
The input image $x$ is fed into SNN encoder $f_e$ which outputs the spiking embedding $x_e \in \{0, 1\}^{d \times T}$, where $d$ is the length dimension and $T$ is the SNN time window. 
Then the latent spiking variable $z_p \in \{0, 1\}^{d \times T}$ is randomly generated by the Poisson process based on the firing rate $r_p \in \left\{\frac{1}{T}, \cdots, \frac{T}{T}\right\}^{d}$ of $x_e$. 
Subsequently, the latent variable $z_p$ is decoded by the SNN decoder $f_d$ and is transformed into the reconstructed image $\hat{x}$. 
During the random image generation process, we first randomly sample a variable $z_n \in \mathbb{R}^d $ from a normal distribution. 
After a bottleneck layer, $z_n$ is converted into the firing rate $r_q \in \left(0, 1\right)^d$ which generate the latent variable $z_q \in \{0, 1\}^{d \times T}$ by Poisson process. 
Finally, the generated image $x'$ is generated by the SNN decoder. 

\subsection{Poisson Based Posterior and Prior Distributions}\label{sec: distribution}
As analyzed in Sec \ref{sec: temporal robustness}, the spiking latent variables show temporal robustness for images generated by the decoder. 
The firing rate of each latent variable neuron has more information relative to the order of the spike firing. 
Therefore, we assume that each neuron of the spiking latent variable $z$ follows a Poisson distribution, which models the number of events in an interval with independent occurrences. 

For a $T$ time window, the probability of emitting $n$ spike during time $T$ follows the Poisson distribution as bellow:
\begin{equation}
    P\left(n \text{ spikes in } T \text{ time steps}\right) = \frac{(r \cdot T)^n}{n!}e^{-r \cdot T},
    \label{eq: probability of Poisson}
\end{equation}
where $r$ means the expectation of the spike firing rate. 
We set $\lambda = r \cdot T$ to denote the intensity parameter of the Poisson distribution, meaning the expected number of spikes in time $T$. 

Then, we denote the posterior probability distribution $p(z_p|x; r_p)$ and the prior probability distribution as $q(z_q; r_q)$, where $r_p$ and $r_q$ are the expectation of the firing rate of posterior and prior, respectively; 
$z_p$ and $z_q$ denote the latent variables generated by posterior and prior, respectively. 

The posterior $p(z_p|x; r_p)$ is modeled by the SNN encoder. 
To project the input image into a latent Poisson distribution space, the firing rate $r_p$ of the encoder output $x_e$ is considered as the expected firing rate of the latent Poisson distribution, with the same length dimension as the spiking latent variable $z_p$. 
For instance, the firing rate $r^i$ of the $i^{th}$ output neuron is computed as:
\begin{equation}
    r_p^i = \frac{1}{T}\sum_{t=1}^{T}x_{e}^{i, t}.
    \label{eq: firing rate}
\end{equation}

For constructing the prior $q(z_q; r_q)$, the distribution of $r_q$ is crucial, as its values encapsulate information of the generated images. 
Therefore, we propose using a bottleneck layer to obtain $r_q$ parameters of the prior.
The bottleneck layer consists of a fully connected layer and a sigmoid active function, as depicted in the generating branch of Fig. \ref{fig: ESVAE}.
The bottleneck input $z_n$ is sampled from a normal distribution, considered the most prevalent naturally occurring distribution. 

\subsection{Reparameterizable Poisson Spiking Sampling}\label{sec: Poisson spiking sampling}
For both the prior and posterior distributions, the target makes the $i^{th}$ neuron of the latent variable fire at a rate $r^i$.
The spike-generating process is modeled as a Poisson process.
Thus, the probability of the $i^{th}$ neuron firing at time $t^{th}$ is expressed as follows:
\begin{equation}
    P\left(\text{Firing at } t^{th} \text{ time step}\right) = r^{i}.
    \label{eq: probability of firing}
\end{equation}

Specifically, we first generate a random variable $u \in \{a|0 \leqslant a \leqslant 1\}^{d \times T}$ from a uniform distribution.
Then, along the time dimension, the value of $u$ is compared with the firing rate $r$ of the corresponding position to generate $z$, which is formulated by:
\begin{equation}
    z^{i, t} = \left\{
    \begin{aligned}
        & 1, u^{i, t} < r^i, \\
        & 0, otherwise.
    \end{aligned}
    \right.
    \label{eq: generate spike}
\end{equation}
where $u^{i, t}$ and $z^{i, t}$ are the $i^{th}$ value at $t^{th}$ time step of $u$ and spiking latent variable $z$.

Since Eq. \ref{eq: generate spike} is a step function, $z$ is not differentiable with respect to $r$. 
To reparameterize $z$, we use the surrogate gradient of our SNN training as follows:
\begin{equation}
    \begin{aligned}
        \frac{\partial z^i}{\partial r^i} &= \sum_{t=1}^{T} \frac{\partial z^{i, t}}{\partial r^i} \\
        &= \frac{1}{\alpha} \sum_{t=1}^{T} \operatorname{sign}\left(|r^i - u^{i, t}| < \frac{\alpha}{2} \right),
    \end{aligned}
\end{equation}
where $\alpha$ is the width parameter to determine the shape of gradient \cite{wu2019direct}.

\subsection{Evidence Lower Bound and Loss Function}\label{sec: ELBO and loss}
The conventional evidence lower bound (ELBO) of VAE is: 
\begin{equation}
    \begin{aligned}
        \operatorname{ELBO} = & \mathbb{E}_{z_p \sim p(z_p|x; r_p)}\left[\log p\left(\hat{x}|z_p\right)\right] \\
        & - \operatorname{KL}\left(p(z_p|x; r_p) || q(z_q; r_q)\right),
    \end{aligned}
    \label{eq: ELBO}
\end{equation}
where $p\left(\hat{x}|z_p\right)$ is the probability distribution function of the reconstructed image $\hat{x}$ generated by $z_p \sim p(z_p|x; r_p)$. 
The first term is usually regarded as the reconstruction loss and reflects the quality of the image reconstructed by $z_p$. 
The second term regularizes the construction of the latent space by reducing the distance between the $p(z_p|x; r_p)$ and $q(z_q; r_q)$ distributions in order to make the model generative. 

Traditional VAEs optimize the KL divergence of these two distributions, and FSVAE argues the MMD metric is more suitable for SNNs \cite{kamata2022fully}. 
However, these metrics are based on the generated spiking latent variables $z_p$ and $z_q$. 
For our ESVAE model, the difference in the spike order of $z_p$ and $z_q$ is not essential, and constraining to reduce their distance will instead make training more difficult.

Therefore, we directly compute the MMD distance between the distribution $p(r_p)$ and $q(r_q)$ of the expected firing rate parameters $r_p$ and $r_q$.
It is formulated by:
\begin{equation}
    \begin{aligned}
        \operatorname{MMD}^2 \left( p(r_p), q(r_q) \right) = & \mathbb{E}_{r_p, r'_p \sim p(r_p)}\left[k \left(r_p, r'_p\right) \right] \\
        & + \mathbb{E}_{r_p, r'_p \sim q(r_q)} \left[k \left(r_q, r'_q\right) \right] \\ 
        & - 2 \mathbb{E}_{r_p \sim p(r_p), r_q \sim q(r_q)} \left[k \left(r_p, r_q\right) \right],
    \end{aligned}
    \label{eq: MMD}
\end{equation}
where $k\left(\cdot, \cdot\right)$ is the kernel function and is set to the radial basis function (RBF) kernel in this paper.

The final loss function $\mathcal{L}$ is derived as follows:
\begin{equation}
    \begin{aligned}
    \mathcal{L} &= \mathbb{E}_{z_p \sim p(z_p|x; r_p)}\left[\log p\left(\hat{x}|z_p\right)\right] \\ 
    &+ \lambda \operatorname{MMD}^2 \left( p(r_p), q(r_q) \right),
    \end{aligned}
    \label{eq: loss}
\end{equation}
where $\lambda$ is a hyperparameter coefficient,
and the empirical estimation of
$\mathbb{E}_{z_p \sim p(z_p|x; r_p)}\left[\log p\left(\hat{x}|z_p\right)\right]$ is calculated by $\operatorname{MSE}(x, \hat{x})$.

\begin{algorithm}[htbp]
    \caption{ESVAE Model Training and Image Generating Algorithms.}
    \label{algorithm}
    \begin{algorithmic}[1]
        \Require Training dataset $\mathcal{X}$.
        \Ensure Reconstructed images $\hat{\mathcal{X}}$, trained SNN encoder $f_e$ and decoder $f_d$, and generated images $\mathcal{X}'$

        \State Initialize the parameters of $f_e$ and $f_d$.
        \While{not done}
            \For {$x$ in $\mathcal{X}$}
                \State $x_e \leftarrow f_e(x)$ 
                \State Calculate the firing rate $r_p$ of $x_e$.  // Eq. \ref{eq: firing rate}
                \State $r_q \leftarrow \Call{Prior}{ }$
                \State $z_p \leftarrow \Call{PoissonProcess}{r_p}$
                \State $\hat{x} \leftarrow f_d(z_p)$
                \State Calculate the loss $\mathcal{L}$ with $x, \hat{x}, r_p, r_q$.  // Eq. \ref{eq: loss}
                \State Update parameters with $\nabla \mathcal{L}$.
            \EndFor
        \EndWhile
        \State $x' \leftarrow \Call{GenerateImages}{ }$
        \State $\mathcal{X}' \leftarrow \mathcal{X}' \cup x'$

        \vspace{3 mm}

        \Function{GenerateImages}{ }
            \State $r_q \leftarrow \Call{Prior}{ }$
            \State $z_q \leftarrow \Call{PoissonProcess}{r_q}$
            \State $x' \leftarrow f_d(z_q)$
            \State \Return{$x'$} 
        \EndFunction

        \Function{Prior}{ }
            \State Randomly sample $z_n$ from $\mathcal{N}(0, 1)$.
            \State $r_q \leftarrow Bottleneck(z_n)$
            \State \Return{$r_q$} 
        \EndFunction

        \Function{PoissonProcess}{$r$}
            \State Randomly sample $u$ from $\mathcal{U}(0, 1)$.
            \State $z \leftarrow \Call{Int}{u < r}$
            \State \Return{$z$}
        \EndFunction

    \end{algorithmic}
\end{algorithm}

The whole model training and image generating process is reported in Algorithm \ref{algorithm}.

\begin{table*}[htbp]
    \centering
    \caption{Performance verification results on different datasets. Our model achieves state-of-the-art performance in most evaluation metrics and has a significant improvement compared with FSVAE. }
    \label{tab: effect verification}
    \renewcommand\arraystretch{1.2}  
    \resizebox{0.95\linewidth}{!}{  
    \begin{tabular}{ccccccc}
    \toprule
    \hline
    \multicolumn{1}{c}{\multirow{2}{*}{Dataset}} & \multicolumn{1}{c}{\multirow{2}{*}{Model}} & \multicolumn{1}{c}{\multirow{2}{*}{Model Type}} & \multicolumn{1}{c}{\multirow{2}{*}{\begin{tabular}[c]{@{}c@{}}Reconstruction \\ Loss $\searrow$ \end{tabular}}} & \multicolumn{1}{c}{\multirow{2}{*}{\begin{tabular}[c]{@{}c@{}}Inception \\ Score $\nearrow$ \end{tabular}}} & \multicolumn{2}{c}{Frechet Distance $\searrow$} \\
    \cline{6-7}
    \multicolumn{1}{c}{} & \multicolumn{1}{c}{} & \multicolumn{1}{c}{} & \multicolumn{1}{c}{} & \multicolumn{1}{c}{} & \multicolumn{1}{c}{Inception (FID)} & \multicolumn{1}{c}{Autoencoder (FAD)} \\ 
    \hline
    \multirow{5}{*}{MNIST} & SWGAN \cite{feng2023spiking} & \multirow{2}{*}{SNN GAN} & - & - & 100.29 & - \\
                           & SGAD \cite{feng2023spiking} &  & - & - & 69.64 & - \\
    \cline{2-7}
                           & ANN \cite{kamata2022fully} & \multirow{3}{*}{SNN VAE} & 0.048 & 5.947 & 112.5 & 17.09 \\
                           & FSVAE \cite{kamata2022fully} & & 0.031 & \textbf{6.209} & \textbf{97.06} & 35.54 \\
                           & \textbf{ESVAE (Ours)} & & \textbf{0.013} & 5.612 & 117.8 & \textbf{10.99} \\
    \hline
    \multirow{5}{*}{
        \begin{tabular}[c]{@{}c@{}}
            Fashion\\ 
            MNIST
        \end{tabular}} & SWGAN \cite{feng2023spiking} & \multirow{2}{*}{SNN GAN} & - & - & 175.34 & - \\
                       & SGAD \cite{feng2023spiking} & & - & - & 165.42 & - \\
    \cline{2-7}
                       & ANN \cite{kamata2022fully} & \multirow{3}{*}{SNN VAE} & 0.050 & 4.252 & 123.7 & 18.08 \\
                       & FSVAE \cite{kamata2022fully} & & 0.031 & 4.551 & \textbf{90.12} & 15.75 \\
                       & \textbf{ESVAE (Ours)} & & \textbf{0.019} & \textbf{6.227} & 125.3 & \textbf{11.13} \\
    \hline
    \multirow{6}{*}{CIFAR10} & SWGAN \cite{feng2023spiking} & \multirow{2}{*}{SNN GAN} & - & - & 178.40 & - \\
                             & SGAD \cite{feng2023spiking} & & - & - & 181.50 & - \\ 
    \cline{2-7}
                             & ANN \cite{kamata2022fully} & \multirow{4}{*}{SNN VAE} & 0.105 & 2.591 & 229.6 & 196.9 \\
                             & FSVAE \cite{kamata2022fully} & & 0.066 & 2.945 & 175.5 & 133.9 \\
                             & TAID \cite{qiu2023when} & & - & 3.53 & 171.1 & 120.5 \\
                             & \textbf{ESVAE (Ours)} & & \textbf{0.045} & \textbf{3.758} & \textbf{127.0} & \textbf{14.74} \\
    \hline
    \multirow{6}{*}{CelebA} & SWGAN \cite{feng2023spiking} & \multirow{2}{*}{SNN GAN} & - & - & 238.42 & - \\
                            & SGAD \cite{feng2023spiking} & & - & - & 151.36 & - \\
    \cline{2-7}
                            & ANN \cite{kamata2022fully} & \multirow{4}{*}{SNN VAE} & 0.059 & 3.231 & 92.53 & 156.9 \\
                            & FSVAE \cite{kamata2022fully} & & 0.051 & 3.697 & 101.6 & 112.9 \\
                            & TAID \cite{qiu2023when} & & - & \textbf{4.31} & 99.54 & 105.3 \\
                            & \textbf{ESVAE (Ours)} & & \textbf{0.034} & 3.868 & \textbf{85.33} & \textbf{51.93} \\  
    \hline
    \bottomrule
    \end{tabular}
    }
\end{table*}

\section{Experiment}
\subsection{Datasets}
MNIST \cite{deng2012mnist} and Fashion MNIST \cite{xiao2017fashion} both have 60,000 training images and 10,000 testing images.
CIFAR10 \cite{krizhevsky2009learning} consists of 50,000 images for training and 10,000 for testing.
For MNIST, Fashion MNIST, and CIFAR10, each image is resized to $32 \times 32$.
CelebA \cite{liu2018large} is a classic face dataset containing 162,770 training samples and 19,962 testing samples.
We resize the images of CelebA to $64 \times 64$.

\subsection{Implementation Details}
\subsubsection{Network Architecture}
Following \cite{kamata2022fully}, we use four conventional layers to construct the backbone of the encoder and decoder on MNIST, Fashion MNIST, and CIFAR10.
The detail of the structure is $32\text{C}3$ - $64\text{C}3$ - $128\text{C}3$ - $256\text{C}3$ - $128\text{FC}$ - $128(\text{sampling})$ - $128\text{FC}$ - $256\text{C}3$ - $128\text{C}3$ - $64\text{C}3$ - $32\text{C}3$ - $32\text{C}3$ - $image\_channel \text{C}3$, where 128 is the latent variable length dimension.
The tdBN \cite{zheng2021going} is inserted in each layer.
For CelebA, we add a $512\text{C}3$ conventional layer in the encoder following $256\text{C}3$ and also in the decoder.
The bottleneck layer comprises a $128\text{FC}$ layer and the Sigmoid active function.

\subsubsection{Training Setting}
For the SNN, we set the time window $T$ to 16, the firing threshold $v_{\theta}$ to 0.2, the membrane potential decay factor $\tau$ to 0.25; the width parameter $\alpha$ of the surrogate gradient to 0.5.
The model is trained 300 epochs by AdamW optimizer with 0.0006 learning rate and 0.001 weight decay.
The learning rate on the bottleneck layer is set to 0.006.

\subsubsection{Hardware Platform}
The source code is written with the Pytorch framework \cite{paszke2019pytorch} on Ubuntu 16.04 LTS. All the models are trained using one NVIDIA GeForce RTX 2080Ti GPU and Intel Xeon Silver 4116 CPU.

\subsection{Performance Verification}  \label{sec: performance}
In this section, we compare our ESVAE with state-of-the-art SNN VAE methods FSVAE (including the ANN version) \cite{kamata2022fully} and TAID \cite{qiu2023when}.
FSVAE is the first fully SNN VAE model which is reported at AAAI22.
TAID adds an attention mechanism based on FSVAE to further improve performance, without the different latent space construction and sampling method, published in ICLR23.
In addition, we also compare the quality of the generated images with the SNN GAN models SWGAN and SGAD \cite{feng2023spiking}.

Table \ref{tab: effect verification} shows the comparison results of different evaluation metrics on each dataset, in which the reconstruction loss, inception score \cite{salimans2016improved} and FID \cite{heusel2017gans} are the commonly used metrics to measure the generated images.
FAD is proposed by \cite{kamata2022fully} to measure the distribution distance between generated and real images.

For the reconstruction loss, our method achieves the lowest loss on both four datasets.
For the generation metrics, the proposed ESVAE also achieves the best results in most items.
It is worth noting that ESVAE gets much better scores on FAD than the other methods.
This means that the posterior distribution $p(z_p|x; r_p)$ constructed explicitly can better project the distribution of the training images.
The experimental results indicate that our method well balances the ability of image restoration and generation.

\begin{figure*}[htbp]
    \centering
	\includegraphics[scale=0.09]{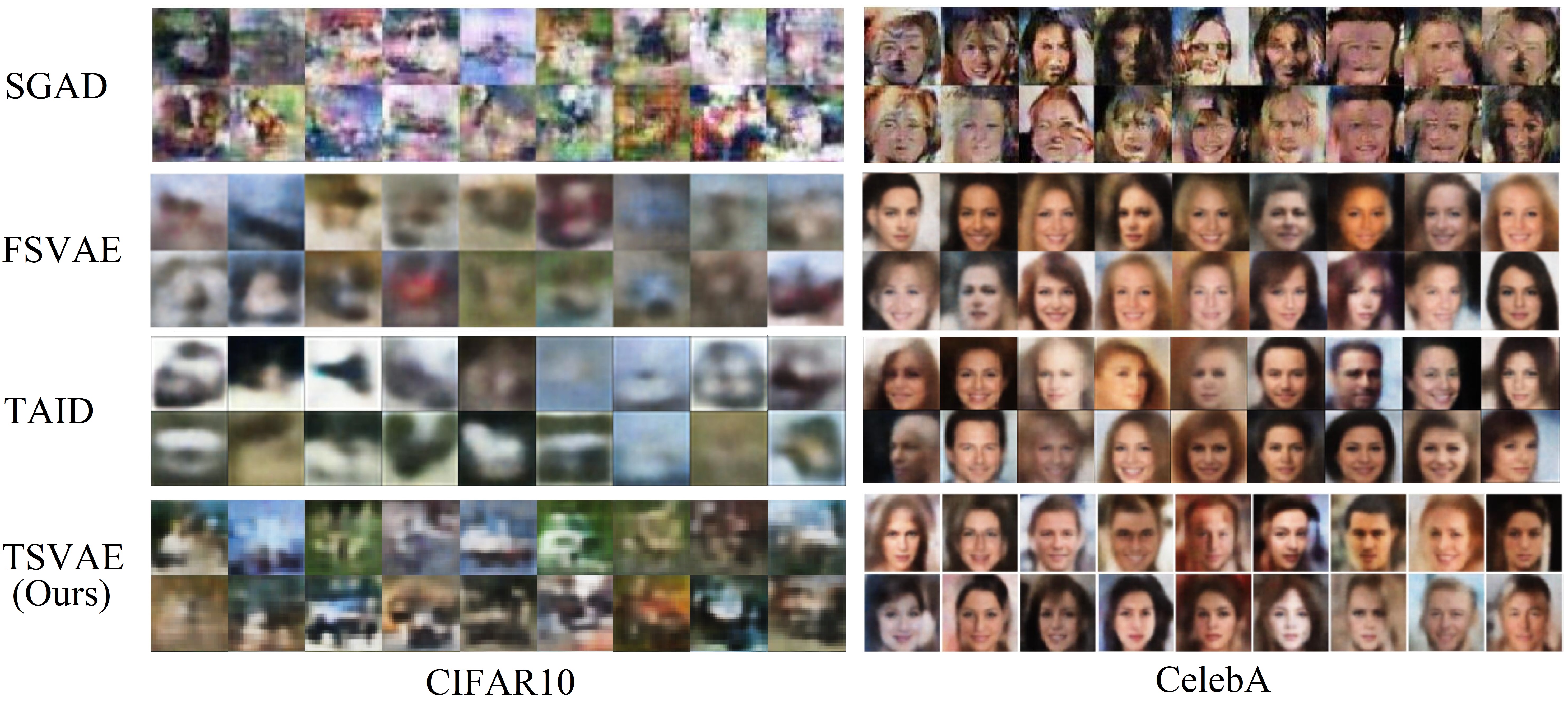}
    \caption{Generated images of SGAD, FSAVE, TAID, and the proposed ESVAE on CIFAR10 and CelebA. }
    \label{fig: generated images}
\end{figure*}

Fig. \ref{fig: generated images} shows the generated images by SGAD, FSVAE, TAID, and ESVAE on CIFAR10 and CelebA. 
Compared with other SNN VAE methods, ESVAE generates images with more details instead of blurred pixels with similar colors.
This is attributed to the better balance between image reduction and generation capabilities in the ESVAE model.
It is worth noting that the images generated by the SNN GAN method SGAD have richer colors and more diverse details.
However, these images lack sufficient rationality, which may be caused by the difficulty of GAN training.

More reconstructed and generated image comparisons are shown in Appendix \ref{appendix: Reconstructed Images} and \ref{appendix: Generated Images}.

\subsection{Robustness Analysis} \label{sec: robustness}
\subsubsection{Temporal Robustness}
We apply the same method as in Section \ref{sec: temporal robustness} to shuffle the latent variables along time dimensions and generate new images.
To compare the temporal robustness accurately, we quantitatively analyze it by calculating the reconstruction loss between images.

\begin{table}[htbp]
    \centering
    \caption{The reconstruction loss of images generated by time-shuffled variables versus original and vanilla reconstructed images.}
    \label{tab: temporal robustness}
    \renewcommand\arraystretch{1.0}  
    \resizebox{1.0\linewidth}{!}{  
    \begin{tabular}{cccc}
    \toprule
    \hline
    Dataset & Model & vs. Original Image & \begin{tabular}[c]{@{}c@{}}vs. Vanilla\\ 
        Reconstructed Image\end{tabular} \\
    
    \hline
    \multirow{2}{*}{MNIST} & FSVAE & 0.0270 & 0.0074 \\
                           & ESVAE & \textbf{0.0105} & \textbf{0.0021} \\
    \hline
    \multirow{2}{*}{\begin{tabular}[c]{@{}l@{}}Fashion\\ MNIST\end{tabular}} & FSVAE & 0.0529 & 0.0265 \\
                    & ESVAE & \textbf{0.0169} & \textbf{0.0030} \\
    \hline
    \multirow{2}{*}{CIFAR10} & FSVAE & 0.0707 & 0.0060 \\
                             & ESVAE & \textbf{0.0434} & \textbf{0.0031} \\
    \hline
    \multirow{2}{*}{CelebA} & FSVAE & 0.0553 & 0.0099 \\
                            & ESVAE & \textbf{0.0330} & \textbf{0.0037} \\
    \hline
    \bottomrule
    \end{tabular}
    }
\end{table}

\begin{figure}[htbp]
    \centering
    \includegraphics[scale=0.55]{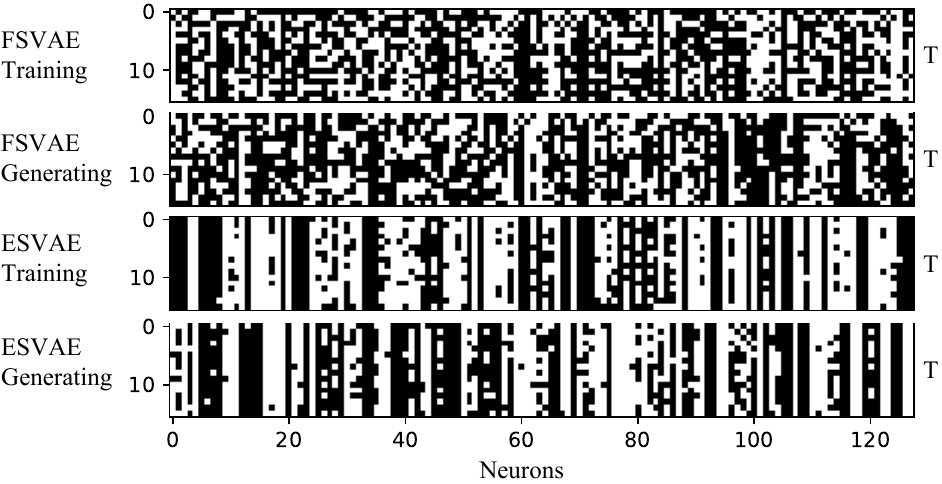}
    \caption{The latent variables sampled by FSVAE and ESVAE on CelebA at the training and generating stage. The horizontal axis is the length dimension of the variable, and the vertical axis is the time dimension.}
    \label{fig: latent variables}
\end{figure}

\begin{figure}[htbp]
    \centering
    \includegraphics[scale=0.57]{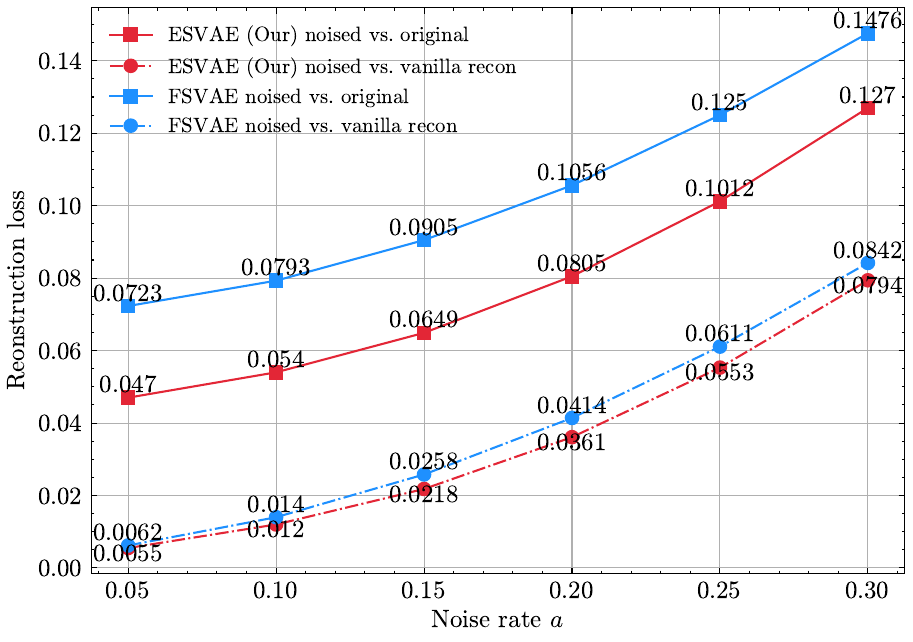}
    \caption{The reconstruction loss curves of noise robustness on CIFAR10. The red lines are the curves of ESVAE, and the blue lines are of FSVAE. Solid and dashed lines show the losses calculated with original and vanilla reconstructed images, respectively.}
    \label{fig: robustness}
\end{figure}

As shown in Table \ref{tab: temporal robustness}, our method has the strongest temporal robustness in comparison with both original and vanilla reconstructed images.
This property makes our method more resistant to problems such as firing delays or hardware confusion.

To further analyze the reason for the better temporal robustness, we visualize the spiking latent variables of CelebA shown in Fig. \ref{fig: latent variables}. 
Observation of the spike trains of each neuron reveals that these neurons have relatively extreme firing characteristics: either high or low firing rates.
This phenomenon is even more pronounced in ESVAE, where many neurons either fire all or not at all.
This feature limits the order of each neuron's spike firing and has a limited influence on the final generated image.
This indicates that the multivariate distribution of the latent space is not obtained by the independent merging of the distributions of different neurons, and the combination order of neurons with different firing rates is also one of the important elements of the distribution of the latent space.

\subsubsection{Gaussian Noise Robustness}
To evaluate the robustness more comprehensively, we add Gaussian noise to the spiking latent variables with probability $a$ on CIFAR10, so that some of the existing spikes disappear or new spikes appear.
As with the test of temporal robustness, we quantify the analysis by calculating the reconstruction loss with the original images and vanilla reconstructed images.
Similarly to the temporal robustness analysis, robustness is quantified by the reconstruction loss between original images and vanilla reconstructed images without noise.

Fig. \ref{fig: robustness} shows the reconstruction loss curves of images generated by noised latent variables.
The experimental results demonstrate that ESVAE is more robust to noise, both in comparison with the original images and with the vanilla reconstructed images.

The images generated by noised latent variables are shown in Appendix \ref{appendix: Noised Images}.

\begin{table*}[htbp]
  \centering
  \caption{Comparison of the amount of computation required to infer a single image in MNIST.}
  \label{tab: power consumption}
  \renewcommand\arraystretch{1.}  
  \resizebox{0.75\linewidth}{!}{  
  \begin{threeparttable}
      \begin{tabular}{ccccc}
          \toprule
          \hline
          \multicolumn{1}{c}{\multirow{2}{*}{Model}} & \multicolumn{2}{l}{Computational complexity} & \multicolumn{1}{c}{\multirow{2}{*}{Average firing rate}} & \multicolumn{1}{c}{\multirow{2}{*}{Power (J)}} \\
          \cline{2-3}
          \multicolumn{1}{c}{} & Addition & Multiplication & \multicolumn{1}{c}{} & \multicolumn{1}{c}{} \\
          \hline
          ANN \cite{kamata2022fully}\tnote{*} & $7.4 \times 10^9$ & $7.4 \times 10^9$ & - & 0.6808 \\
          FSVAE \cite{kamata2022fully} & $5.0 \times 10^{10}$ & $5.6 \times 10^8$ & \textbf{0.3390} & 0.2468 \\
          \textbf{ESVAE (Ours)} & $\mathbf{1.9 \times 10^8}$ & $\mathbf{1.8 \times 10^6}$ & 0.4491 & \textbf{0.0012} \\
          \hline
          \bottomrule
  \end{tabular}
  \begin{tablenotes}    
    \footnotesize               
    \item[*] The ANN VAE method is introduced in ref \cite{kamata2022fully}, and has the same backbone as ESVAE and FSVAE.  
  \end{tablenotes}  
  \end{threeparttable}
  }
\end{table*}

\begin{figure*}[htbp]
    \centering
    \begin{subfigure}[b]{0.34\textwidth}
        \includegraphics[width=\textwidth]{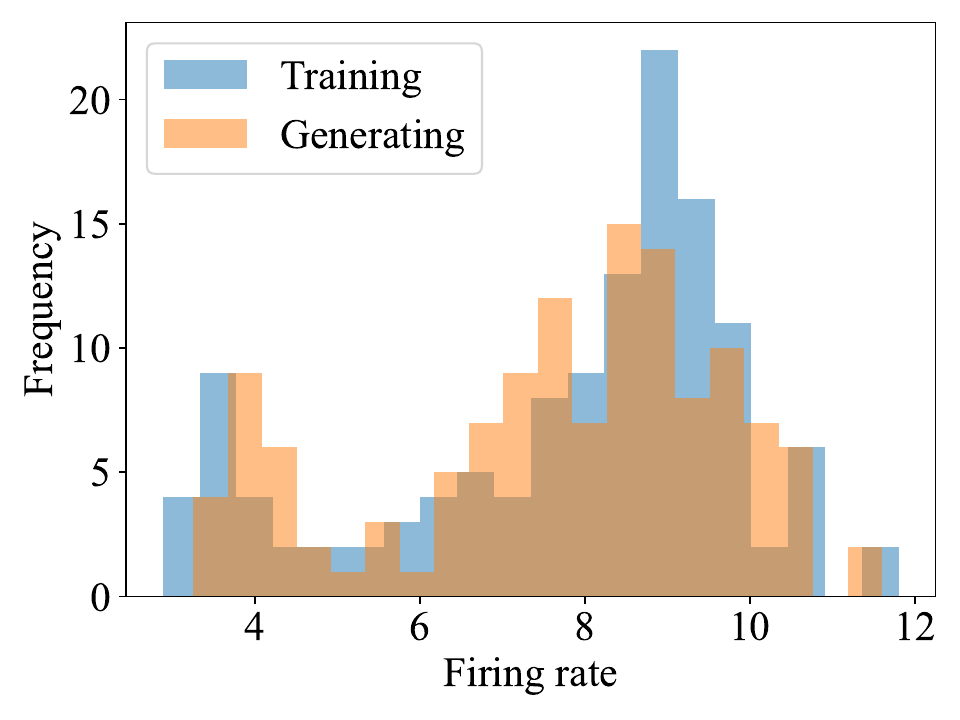}
        \caption{Mean firing rates of FSVAE.}
        \label{fig: FSVAE mean firing rate}
    \end{subfigure}
    \begin{subfigure}[b]{0.34\textwidth}
        \includegraphics[width=\textwidth]{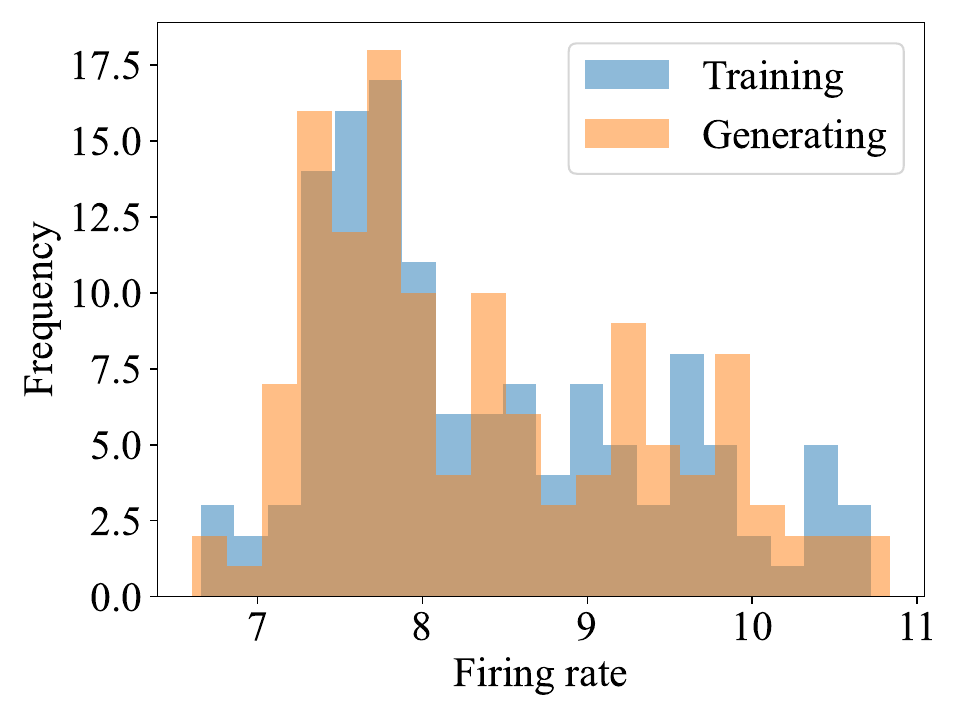}
        \caption{Mean firing rates of ESVAE.}
        \label{fig: ESVAE mean firing rate}
    \end{subfigure}
    \begin{subfigure}[b]{0.34\textwidth}
        \includegraphics[width=\textwidth]{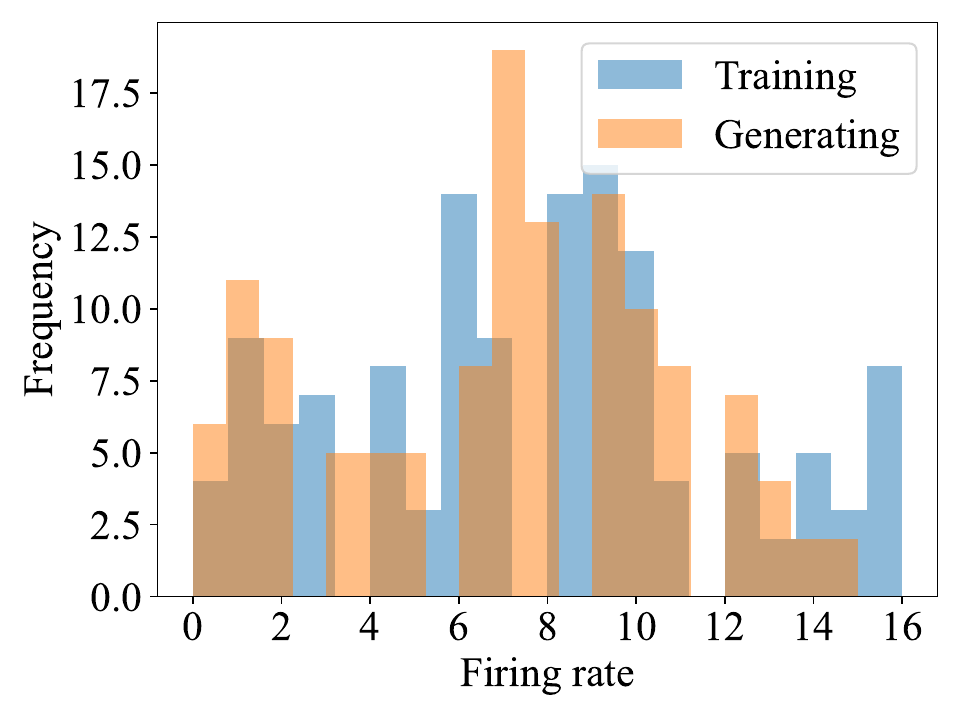}
        \caption{Single sample firing rates of FSVAE.}
        \label{fig: FSVAE firing rate of one sample}
    \end{subfigure}
    \begin{subfigure}[b]{0.34\textwidth}
        \includegraphics[width=\textwidth]{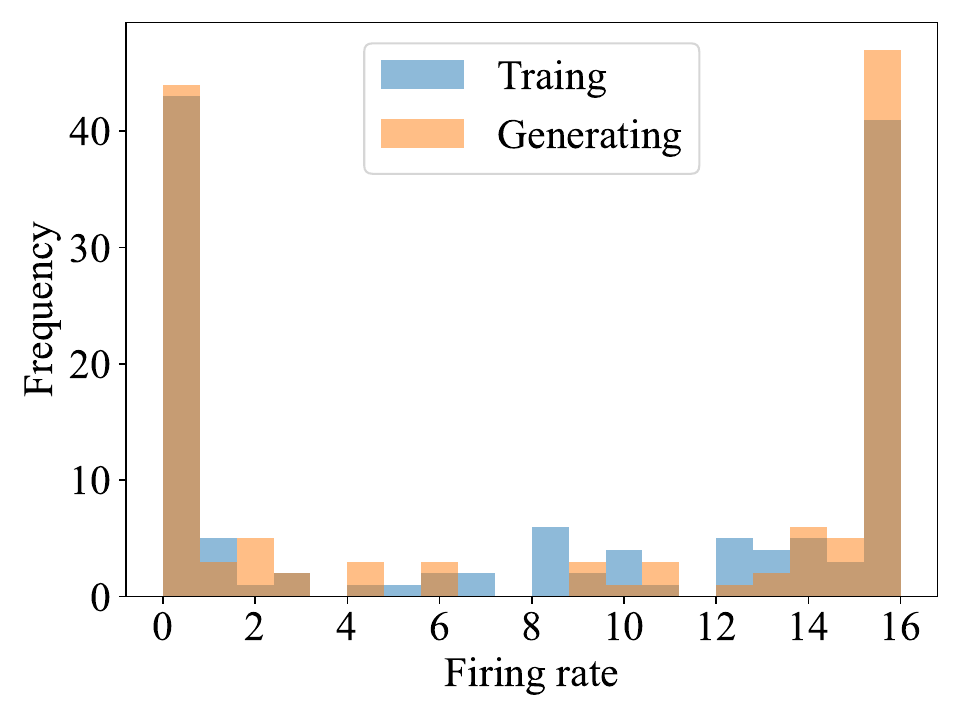}
        \caption{Single sample firing rates of ESVAE.}
        \label{fig: ESVAE firing rate of one sample}
    \end{subfigure}
    \caption{
        The firing rate distribution of FSVAE and ESVAE on the CIFAR10 dataset. 
        Fig. \ref{fig: FSVAE mean firing rate} and \ref{fig: ESVAE mean firing rate} are the mean firing rate distribution of all the generated images. 
        Fig. \ref{fig: FSVAE firing rate of one sample} and \ref{fig: ESVAE firing rate of one sample} are the firing rate distribution of a single image generated by FSVAE and ESVAE, respectively. 
    }
\label{fig: firing rate}
\end{figure*}

\subsection{Comparison on Energy Efficiency}
In this section, we compare the computation cost and energy efficiency of our ESVAE and baseline methods.
Specifically, we count the number of floating-point addition and multiplication operations required by the inference process to generate an image on MNIST under the same structure.
The synaptic operations (SOPs) in SNN can be calculated as follows (\cite{zhou2023spikformer}):
\begin{equation}
    \operatorname{SOPs} = \bar{r} \cdot T \cdot \operatorname{FLOPs},
\end{equation}
where $\operatorname{FLOPs}$ represents the sum numbers of addition and multiplication floating-point operations, $\bar{r}$ is the average firing rate of the SNN model.
Following \cite{zhou2023spikformer, horowitz20141}, the floating-point and synaptic operations consume 4.6$p$J and 0.9$p$J of energy, respectively.

Table \ref{tab: power consumption} shows the comparison result.
Our ESVAE method directly eliminates the extra sampling network of the baseline FSVAE on the same backbone, reducing a large number of calculations.
Although the FSVAE method consumes less power than the ANN, this is due to its low firing rate and low power consumption for synaptic operations. 
The FSVAE method, instead, has a higher amount of addition calculations than the ANN.
Our ESVAE method achieves the maximum advantage in both additive and multiplicative computations compared to FSVAE.
Even though the ESVAE method has a higher firing rate than FSVAE, the advantage in terms of computational complexity allows ESVAE to obtain a cross-order of magnitude reduction in power consumption.

\begin{figure*}[htbp]
    \centering
    \begin{subfigure}[b]{0.34\textwidth}
        \includegraphics[width=\textwidth]{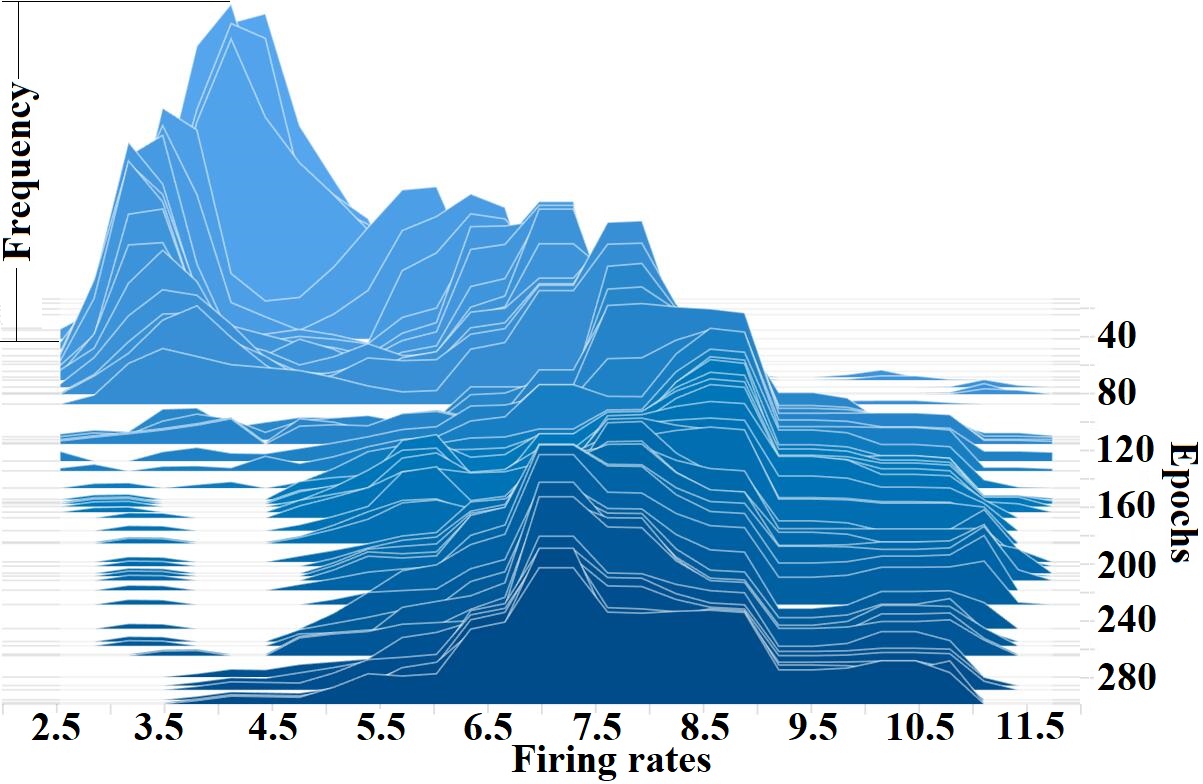}
        \caption{FSVAE training.}
        \label{fig: FSVAE training distribution}
    \end{subfigure}
    \begin{subfigure}[b]{0.34\textwidth}
        \includegraphics[width=\textwidth]{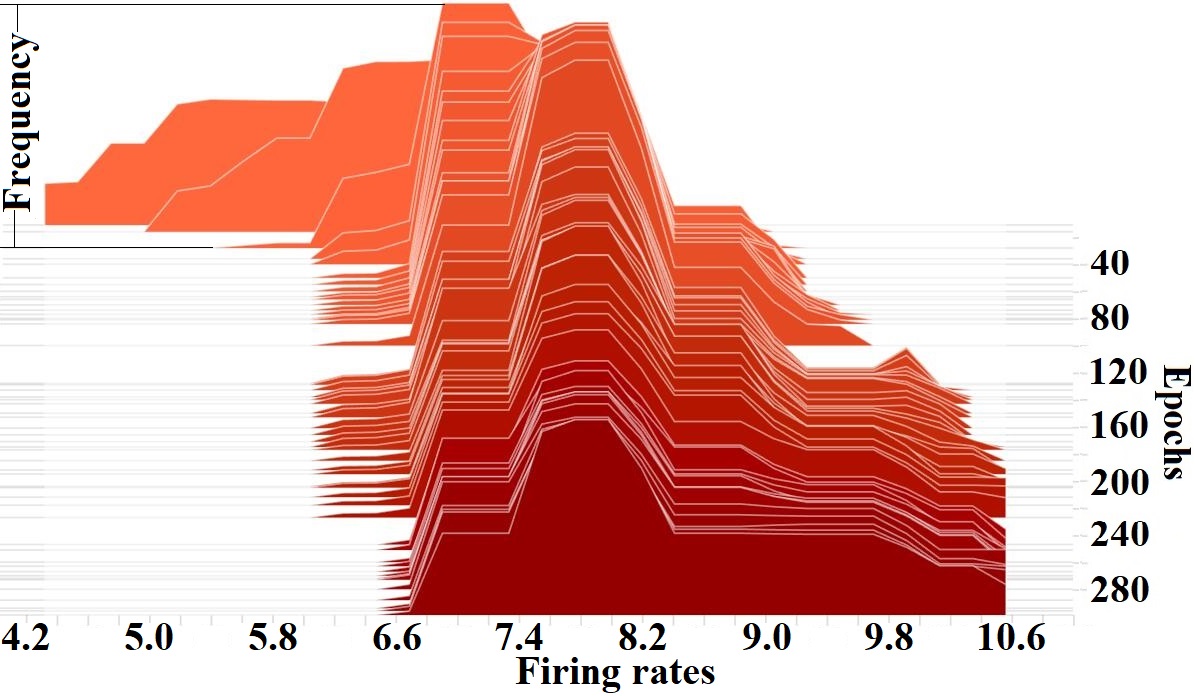}
        \caption{ESVAE training.}
        \label{fig: ESVAE training distribution}
    \end{subfigure}
    \begin{subfigure}[b]{0.34\textwidth}
        \includegraphics[width=\textwidth]{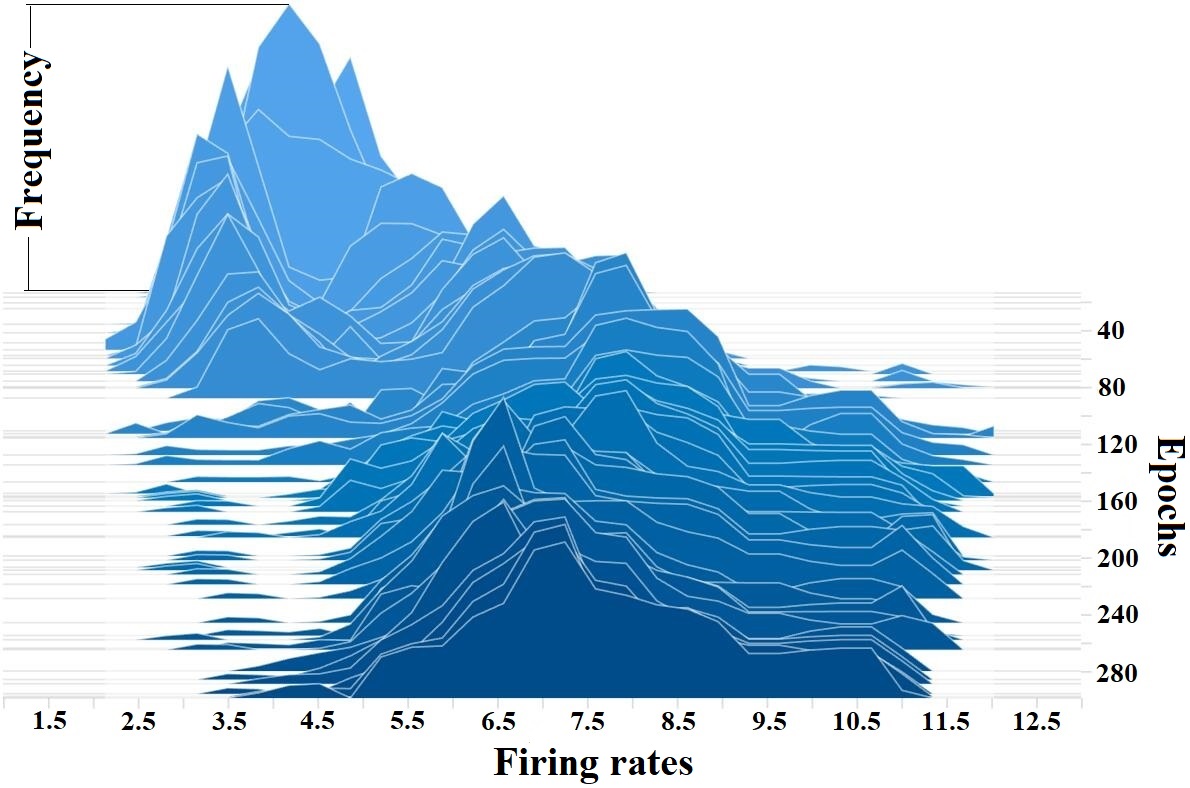}
        \caption{FSVAE generating.}
        \label{fig: FSVAE sampling distribution}
    \end{subfigure}
    \begin{subfigure}[b]{0.34\textwidth}
        \includegraphics[width=\textwidth]{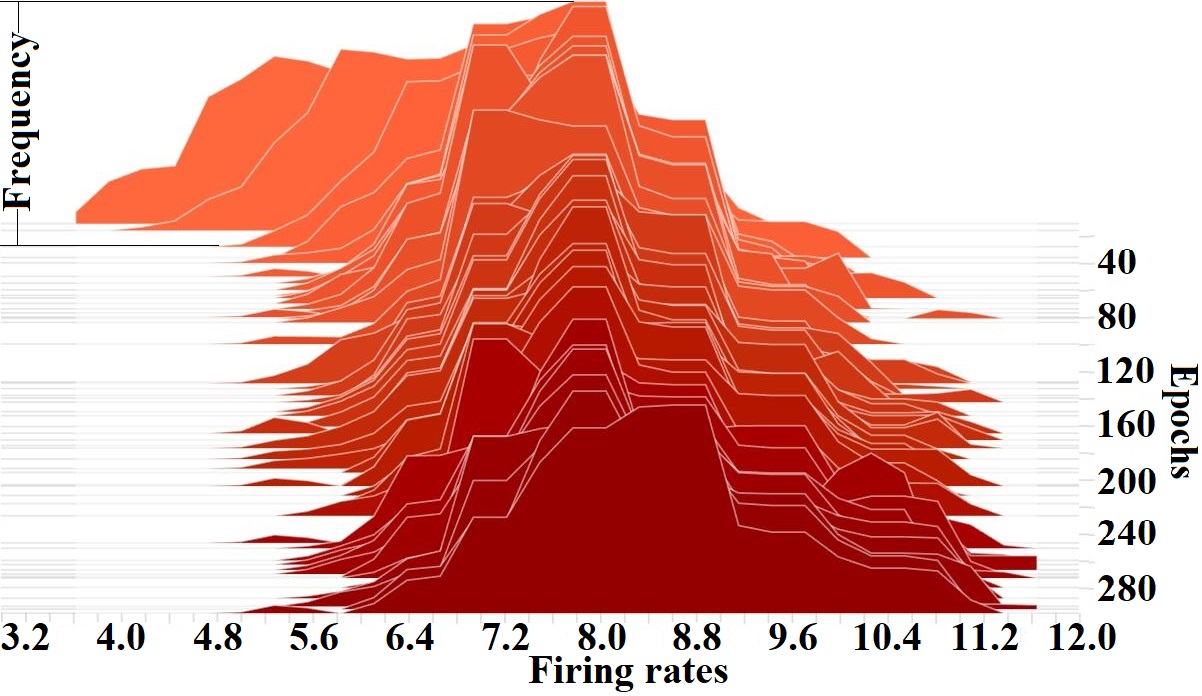}
        \caption{ESVAE generating.}
        \label{fig: ESVAE sampling distribution}
    \end{subfigure}
    \caption{
        Firing rates distributions of FSVAE and ESVAE at different training epochs on CIFAR10. The horizontal axis represents the firing rate, while the vertical axis depicts frequency. Different colors represent different training epochs.
    }
    \label{fig: training and sampling distributions}
\end{figure*}

\begin{table}[hbp]
    \centering
    \caption{The classification accuracies (\%) of encoder output.}
    \label{tab: Classification}
    \renewcommand\arraystretch{1.0}  
    \resizebox{0.8\linewidth}{!}{  
    \begin{tabular}{ccccc}
    \toprule
    \hline
     & CIFAR10 & Fashion MNIST & MNIST\\ 
    \hline
    FSVAE & 46.65 & 86.28 & 98.00 \\
    ESVAE & \textbf{53.59} & \textbf{88.59} & \textbf{98.09} \\
    \hline
    \bottomrule
    \end{tabular}
    }
\end{table}

\subsection{Comparison of Encoder on Classification Task}
The powerful image generation capabilities and robustness of ESVAE demonstrated in Sec. \ref{sec: performance} and \ref{sec: robustness} are mainly brought by the SNN decoder. 
We now analyze the capabilities of the encoder by classifying the encoder embeddings $x_e$.
We feed the firing rate $r_p$ of $x_e$ into an ANN classifier which consists of four fully connected layers: $128$-$512\text{FC}$-$256\text{FC}$-$128\text{FC}$-$10\text{FC}$.

The classifier is trained 200 epochs by an SGD optimizer with a 0.01 learning rate on the training set.
Table \ref{tab: Classification} shows the test accuracy on CIFAR10, Fashion MNIST, and MNIST.
The results show that ESVAE achieves the highest classification accuracy on all datasets.
Especially on CIFAR10, the accuracy of ESVAE is 6.94\% higher than FSVAE, which indicates that our encoder preserves more information about the input image on the complex dataset.

\subsection{Comparison on Distribution Consistency}
In this section, we analyze the posterior and prior distribution consistency between training and generating stages, by visualizing the frequency of the firing rate in all neurons of latent variables.
The visualization results are shown in Fig. \ref{fig: firing rate}.
For the distribution of mean firing rates, the overlap between the training and generating stages is high on both ESVAE Fig. \ref{fig: ESVAE mean firing rate} and FSVAE Fig. \ref{fig: FSVAE mean firing rate}. 
This suggests that the distance of the distributions can also be effectively reduced by optimizing the MMD loss between firing rates.

Another interesting observation is that the distribution of a single generated image, as shown in Fig. \ref{fig: ESVAE firing rate of one sample}, is far from the mean firing rate distribution.
The distribution shown in Fig. \ref{fig: FSVAE firing rate of one sample} is also not the same as the mean firing rate distribution of FSVAE, but the gap is smaller than that of ESVAE.
We believe that this difference is brought about by the distinction between the different categories of samples. 
In ESVAE, the difference between the distribution of individual samples and the mean distribution is greater, which also suggests a higher distinction between the samples. 
The better classification ability shown in Table \ref{tab: Classification} can also prove this conclusion. 

Fig. \ref{fig: training and sampling distributions} shows the distribution changes of FSVAE and ESVAE during training and generating processes. 
The firing rate distribution of ESVAE training converges faster than FSVAE.
The distribution shape of ESVAE training remains consistent mainly in early training, with the generating distribution closely matching it.
The distribution of FSVAE generating is also highly consistent with the training distribution, yet it takes more training epochs to converge to a stable distribution shape.

\section{Discussion with Other Spiking Generation Models}
Recently, many achievements have been made in the field of image generation with SNNs, such as the SNN-based diffusion model SDDPM \cite{cao2024spiking} and Spiking VQ-VAE \cite{feng2024time}.
Our work is not intended to compete with these methods in terms of image fidelity.
Instead, we propose a new SNN VAE method that can be applied to other methods with spiking prior and posterior presentation and stochastic sampling process.
The innovative spiking sampling mechanism of our ESVAE opens up intriguing possibilities for theoretical integration with other VAE-based methods, such as the diffusion model. 

Since SNN methods such as SDDPM and Spiking VQ-VAE, which can generate high-definition images, require huge computational resources, e.g., 4 A100 GPUs, we analyze the applicability of ESVAE on other traditional SNN VAE applications.

TAID and PFA (Projected-full Attention) are two spiking attention methods applied to the spiking VAE task on references \cite{qiu2023when} and \cite{deng2024tensor}, respectively.
We validate the applicability of our method by replacing the backbone VAE model in these two methods with the proposed ESVAE model.

\begin{table}[htbp]
    \centering
    \caption{Applicability verification results of ESVAE on CIFAR10. }
    \label{tab: applicability verification}
    \renewcommand\arraystretch{1.2}  
    \resizebox{1.0\linewidth}{!}{  
    \begin{tabular}{cccc}
    \toprule
    \hline
    \multicolumn{1}{c}{\multirow{2}{*}{Model}} & \multicolumn{1}{c}{\multirow{2}{*}{\begin{tabular}[c]{@{}c@{}}Inception \\ Score $\nearrow$ \end{tabular}}} & \multicolumn{2}{c}{Frechet Distance $\searrow$} \\
    \cline{3-4}
    \multicolumn{1}{c}{} & \multicolumn{1}{c}{} & \multicolumn{1}{c}{Inception (FID)} & \multicolumn{1}{c}{Autoencoder (FAD)} \\ 
    \hline
    TAID \cite{qiu2023when} & 3.53 & 171.1 & 120.5 \\
    \textbf{ESVAE+TAID} & \textbf{3.719} & \textbf{130.7} & \textbf{18.32} \\
    \hdashline
    PFA \cite{deng2024tensor} & \textbf{3.84} & 166.4 & 92.83 \\
    \textbf{ESVAE+PFA} & 3.655 & \textbf{135.4} & \textbf{19.10} \\
    \hline
    \bottomrule
    \end{tabular}
    }
\end{table}

Table \ref{tab: applicability verification} shows the verification results on CIFAR10.
The proposed ESVAE has also achieved a significant improvement in the effectiveness of these two methods.
The experimental results demonstrate that ESVAE has the potential to act as a superior backbone model among other VAE-related methods.

\section{Conclusion}
In this paper, we propose an ESVAE model with a reparameterizable Poisson spiking sampling method.
The latent space in SNN is explicitly constructed by a Poisson-based posterior and prior distribution, which improves the interpretability and performance of the model.
Subsequently, the proposed sampling method is used to generate the spiking latent variables by the Poisson process, and the surrogate gradient mechanism is introduced to reparameterize the sampling method.
We conduct comprehensive experiments on benchmark datasets.
Experimental results show that images generated by ESVAE outperform the existing SNN VAE and SNN GAN models.
Moreover, ESVAE has stronger robustness, a higher distribution consistency, and an encoder with more information retention.

\textbf{Limitation and future works:} 
While the proposed ESVAE demonstrates promising computational efficiency and spiking sample mechanism, it currently falls short in generating high-fidelity images compared to state-of-the-art diffusion models. 
In future work, we will try to incorporate our ESVAE method into other state-of-the-art SNN image generation methods and further reduce their training costs. 
Additionally, we will explore new SNN VAE methods, which are suitable for event data recorded by neuromorphic cameras, to expand the practice of SNNs.


{\appendices
\section{Reconstructed Images} \label{appendix: Reconstructed Images} 
Fig. \ref{fig: reconstructed images on CIFAR10} and \ref{fig: reconstructed images on CelebA} compare the reconstructed images of FSVAE \cite{kamata2022fully} and ESVAE on CIFAR10 and CelebA, respectively.
Our ESVAE demonstrates a higher-quality reconstruction with more image detail than the fuzzy blocks of color demonstrated by FSVAE. 

\section{Generated Images} \label{appendix: Generated Images}
The randomly generated images of FSVAE and ESVAE are shown in Fig. \ref{fig: generated images on CIFAR10} and \ref{fig: generated images on CelebA}, respectively. 
As the reconstructed images, the generated images of ESVAE have richer color variations.

\section{Noised Images} \label{appendix: Noised Images}
Fig. \ref{fig: noised images on CIFAR10} and \ref{fig: noised images on CelebA} illustrate the images generated by the latent variables with different noise disturbances on CIFAR10 and CelebA, respectively.
For ESVAE, disrupting the spike order of the latent variables, the generated images are almost no different from the vanilla reconstructed images. 
Under Gaussian noise interference, it can be seen that when $a$ reaches 0.1, the images generated by FSVAE are more different from the variables reconstructed images, while ESVAE still maintains the distinctive features of the original images to a greater extent.
\begin{figure*}[ht]
    \centering
    \includegraphics[scale=1.0]{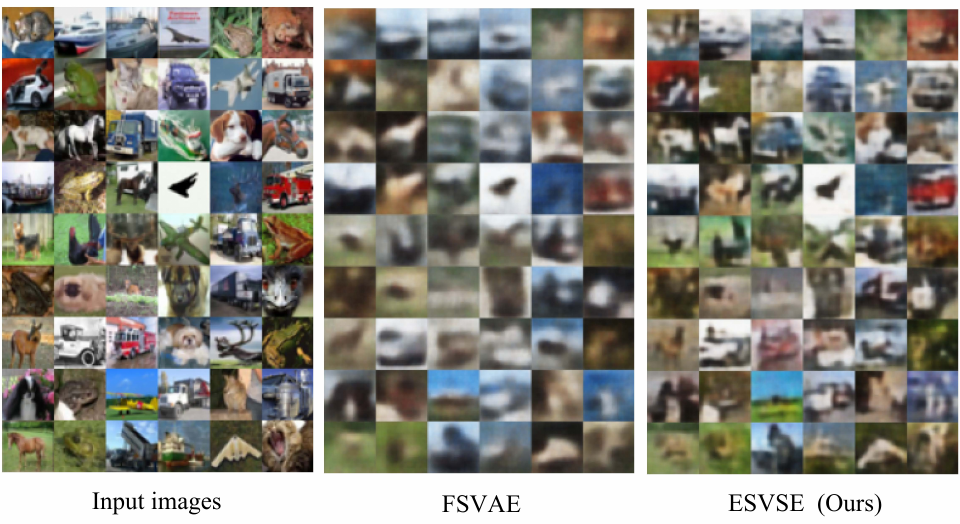}
    \caption{Reconstructed images of FSAVE and ESVAE on CIFAR10. }
    \label{fig: reconstructed images on CIFAR10}
\end{figure*}
\begin{figure*}[ht]
    \centering
	\includegraphics[scale=1.0]{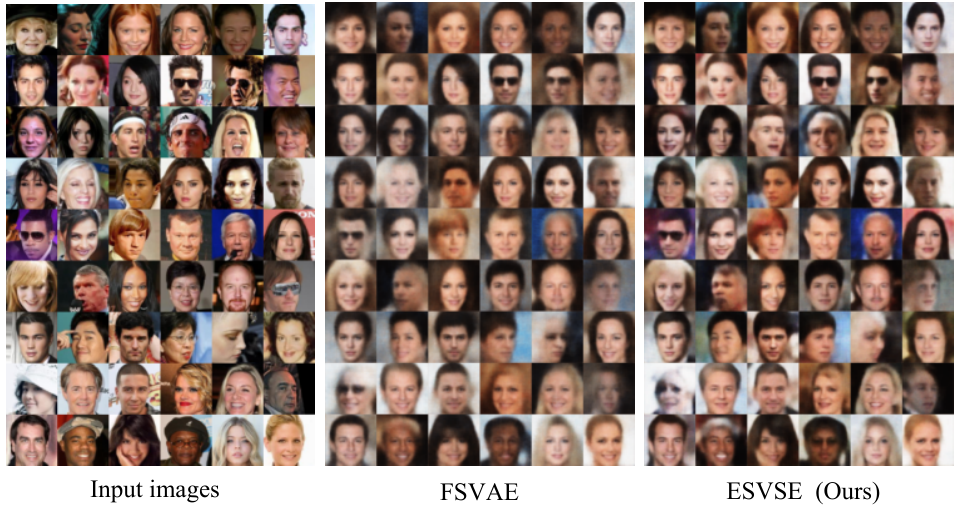}
    \caption{Reconstructed images of FSAVE and ESVAE on CelebA. }
    \label{fig: reconstructed images on CelebA}
\end{figure*}
\begin{figure*}[ht]
    \centering
	\includegraphics[scale=0.95]{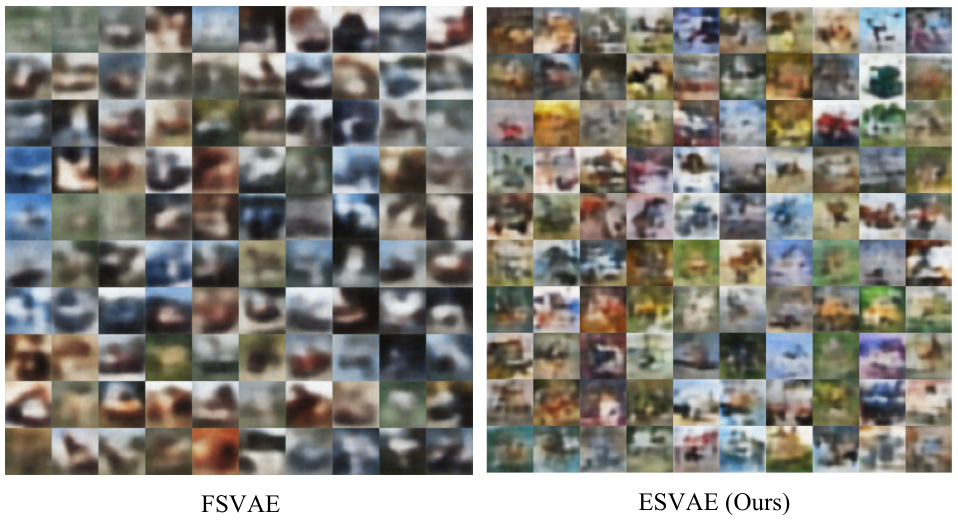}
    \caption{Generated images of FSAVE and ESVAE on CIFAR10. }
    \label{fig: generated images on CIFAR10}
\end{figure*}
\begin{figure*}[ht]
    \centering
	\includegraphics[scale=0.95]{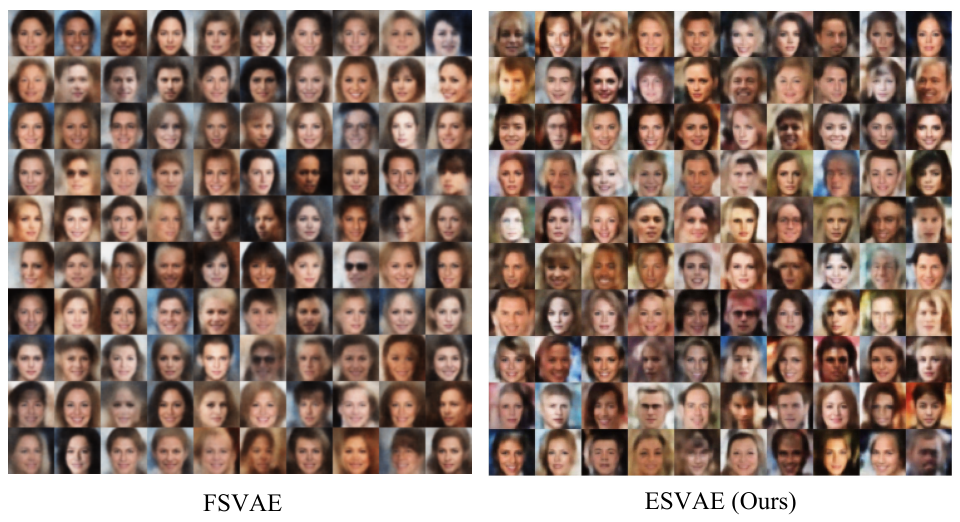}
    \caption{Generated images of FSAVE and ESVAE on CelebA. }
    \label{fig: generated images on CelebA}
\end{figure*}
\begin{figure*}[ht]
    \centering
	\includegraphics[scale=0.8]{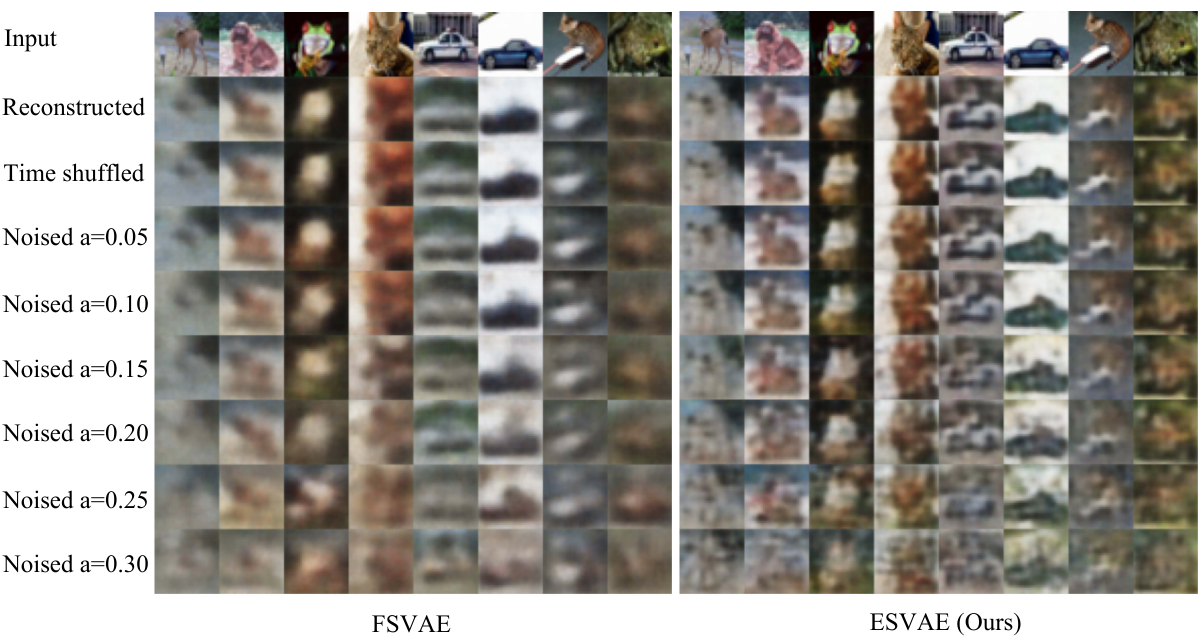}
    \caption{Noised images of FSAVE and ESVAE on CIFAR10. }
    \label{fig: noised images on CIFAR10}
\end{figure*}
\begin{figure*}[ht]
    \centering
	\includegraphics[scale=0.8]{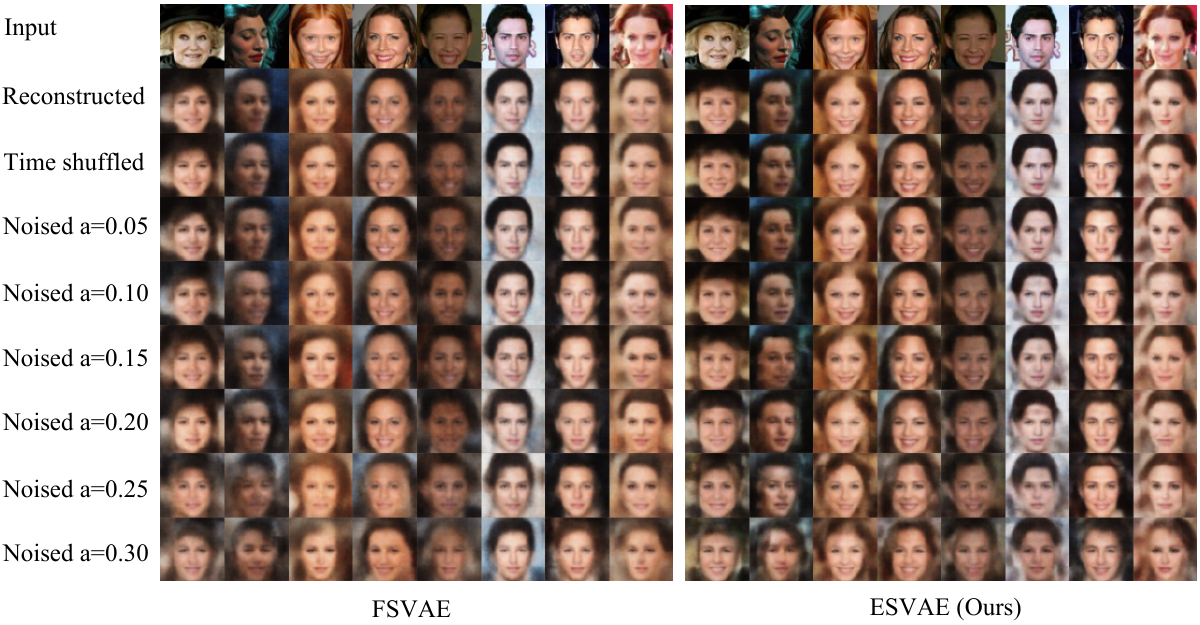}
    \caption{Noised images of FSAVE and ESVAE on CelebA. }
    \label{fig: noised images on CelebA}
\end{figure*}
}



\bibliographystyle{IEEEtran}
\bibliography{bare_jrnl_new_sample4}

\begin{IEEEbiography}[{\includegraphics[width=1in,height=1.25in,clip,keepaspectratio]{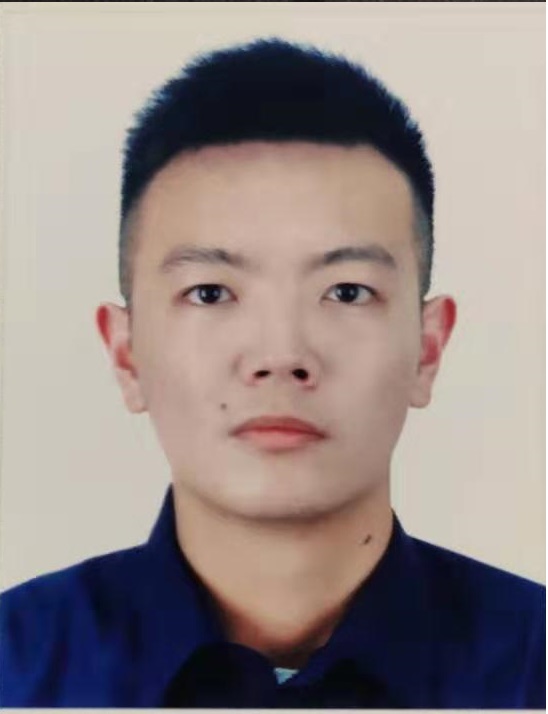}}]{Qiugang Zhan} received his B.S. degree from Yantai University, Yantai, China, in 2017 and his Ph.D. degree in computer science from the University of Electronic Science and Technology of China, Chengdu, China, in 2024.
He is currently a lecturer with the School of Computing and Artificial Intelligence, Southwestern University of Finance and Economics, Chengdu. 
His research interests include spiking neural networks, federated learning and transfer learning.
\end{IEEEbiography}

\begin{IEEEbiography}[{\includegraphics[width=1in,height=1.25in,clip,keepaspectratio]{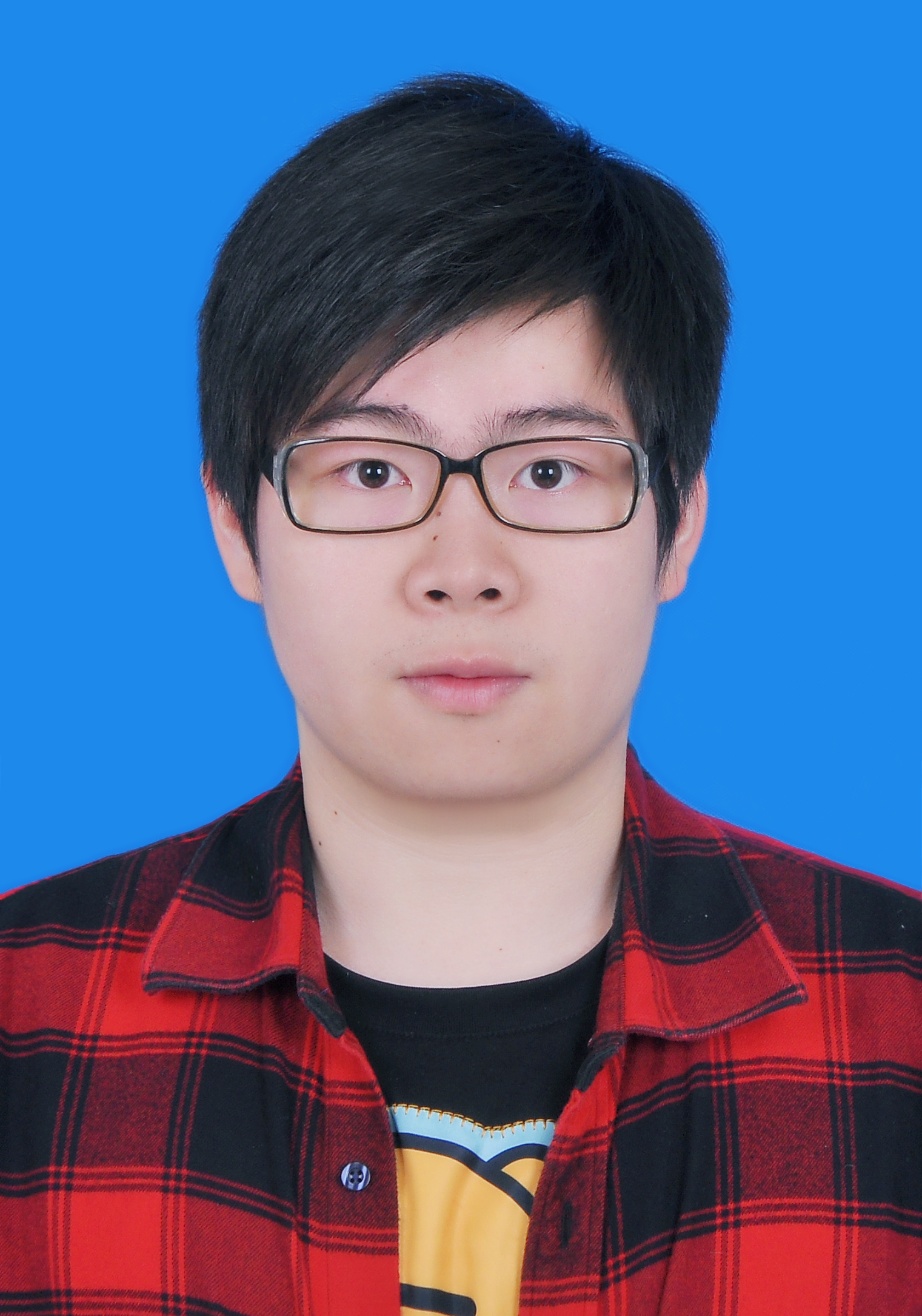}}]{Ran Tao} received the bachelor's degree from Southwest University, Chongqing, China, in 2020, and the master's degree from the University of Auckland, Auckland, New Zealand, in 2021. He is currently pursuing for the Ph.D. degree with the School of Computing and Artificial Intelligence, Southwestern University of Finance and Economics. His current research interests include federated learning and spiking neural networks.
\end{IEEEbiography}

\begin{IEEEbiography}[{\includegraphics[width=1in,height=1.25in,clip,keepaspectratio]{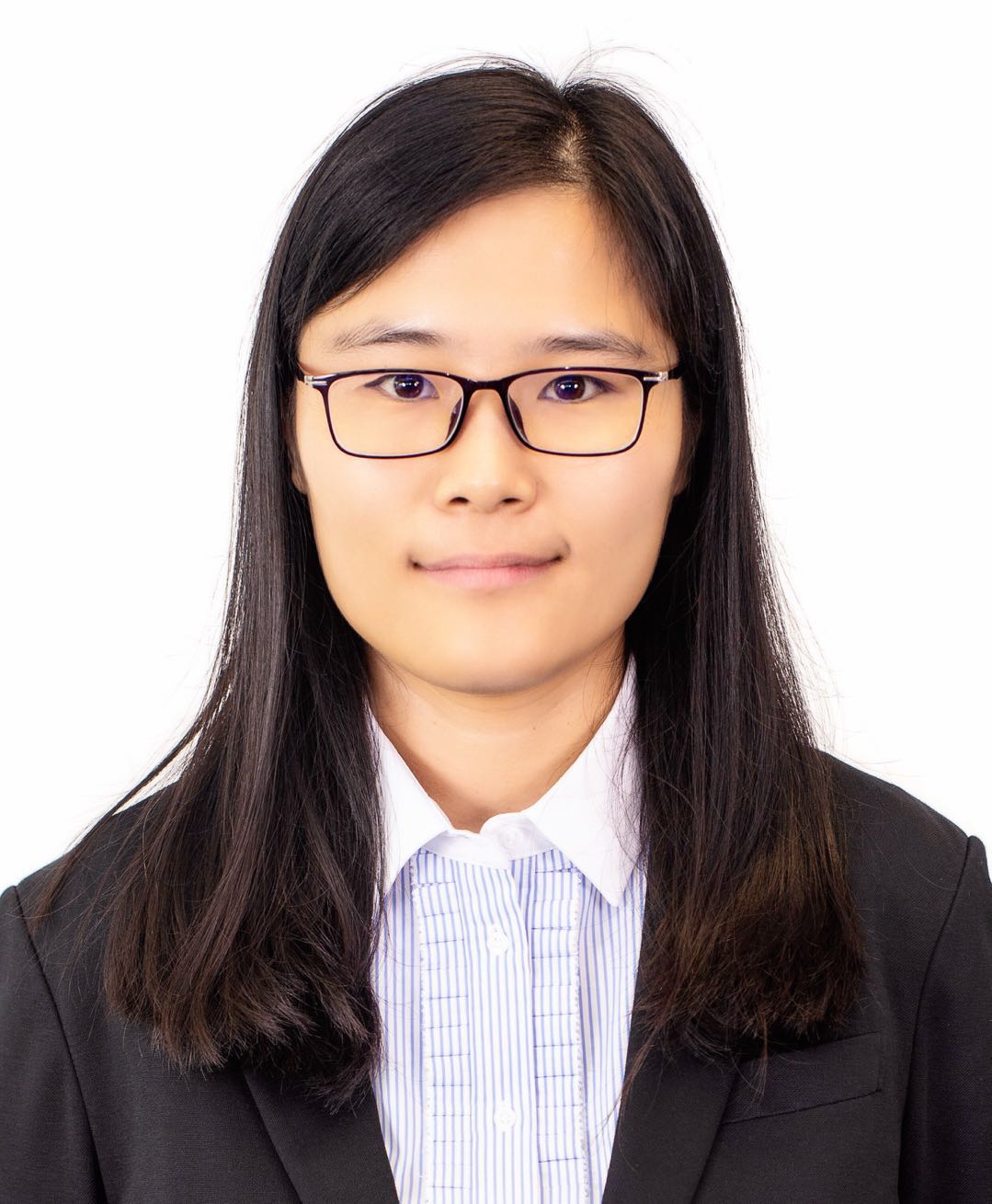}}]{Xiurui Xie} received the Ph.D. degree in computer science from the University of Electronic Science and Technology of China, Chengdu, China, in 2016. Dr. Xie worked as a Research Fellow in Nanyang Technological University, Singapore from 2017 to 2018, and worked as a Research Scientist in the Agency for Science, Technology and Research (ASTAR), Singapore from 2018 to 2020. She has authored over 10 technical papers in prominent journals and conferences. Her primary research interests are neural networks, neuromorphic chips, transfer learning and pattern recognition.
\end{IEEEbiography}

\begin{IEEEbiography}[{\includegraphics[width=1in,height=1.25in,clip,keepaspectratio]{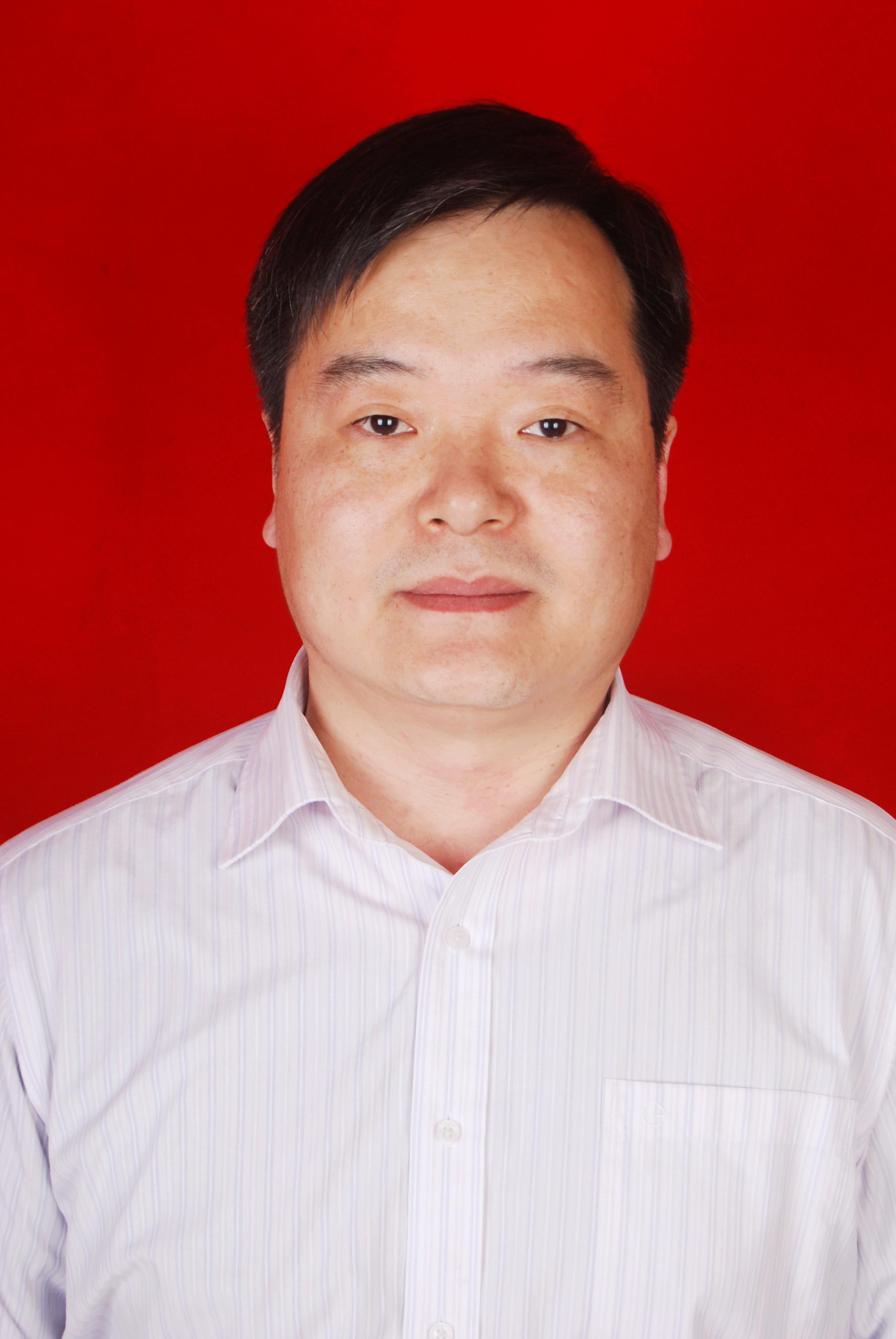}}]{Guisong Liu} (Member, IEEE) received the B.S. degree in mechanics from Xi’an Jiao Tong University, Xi’an, China, in 1995, and the M.S. degree in automatics and the Ph.D. degree in computer science from the University of Electronic Science and Technology of China, Chengdu, China, in 2000 and 2007, respectively. 
    
He was a Visiting Scholar with Humbolt University, Berlin, Germany, in 2015. Before 2021, he was a Professor with the School of Computer Science and Engineering, the University of Electronic Science and Technology of China. He is currently a Professor and the Dean of the School of Computing and Artificial Intelligence, Southwestern University of Finance and Economics, Chengdu. He has filed over 20 patents, and published over 70 scientific conference and journal papers. His research interests include pattern recognition, neural networks, and machine
learning.
\end{IEEEbiography}

\begin{IEEEbiography}[{\includegraphics[width=1in,height=1.25in,clip,keepaspectratio]{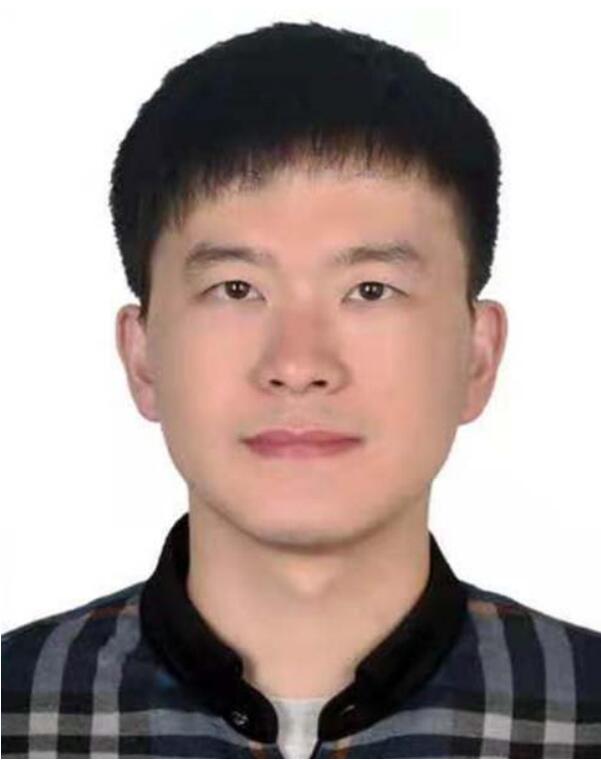}}]{Malu Zhang} (Member, IEEE) received the Ph.D. degree in computer science from the University of Electronic Science and Technology of China, Chengdu, China, in 2019.
From 2019 to 2022, he was a Research Fellow with the HLT Laboratory, Department of Electrical and Computer Engineering, National University of Singapore, Singapore. He is currently a Professor with the School of Computer Science and Engineering, University of Electronic Science and Technology of China. His research interests include spiking neural networks, neural spike encoding, and neuromorphic applications. Dr. Zhang is now an Associate Editor of the IEEE Transactions on Emerging Topics in Computational Intelligence.
\end{IEEEbiography}

\begin{IEEEbiography}[{\includegraphics[width=1in,height=1.25in,clip,keepaspectratio]{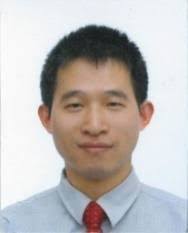}}]{Huajin Tang} (Senior Member, IEEE) received the B.Eng. degree from Zhejiang University, China in 1998, received the M.Eng. degree from Shanghai Jiao Tong University, China in 2001, and received the Ph.D. degree from the National University of Singapore, in 2005. 
    
He was a system engineer with STMicroelectronics, Singapore, from 2004 to 2006. From 2006 to 2008, he was a Post-Doctoral Fellow with the Queensland Brain Institute, University of Queensland, Australia. Since 2008, he was Head of the Robotic Cognition Lab, Institute for Infocomm Research, A*STAR, Singapore. Since 2014 he is a Professor with College of Computer Science, Sichuan	University and now he is a Professor with College of Computer Science and Technology, Zhejiang University, China. He received the 2016 IEEE Outstanding TNNLS Paper Award and 2019 IEEE Computational Intelligence Magazine Outstanding Paper Award. His current research interests include neuromorphic computing, neuromorphic hardware and cognitive systems, robotic cognition, etc.
    
Dr. Tang has served as an Associate Editor of IEEE Trans. on Neural Networks and Learning Systems, IEEE Trans. on Cognitive and Developmental Systems, Frontiers in Neuromorphic Engineering, and Neural Networks. He was the Program Chair of IEEE CIS-RAM (2015, 2017), and ISNN (2019), and Co-Chair of IEEE Symposium on Neuromorphic Cognitive Computing (2016-2019). He is a Board of Governor member of International Neural Networks Society.
\end{IEEEbiography}

\begin{IEEEbiography}[{\includegraphics[width=1in,height=1.25in,clip,keepaspectratio]{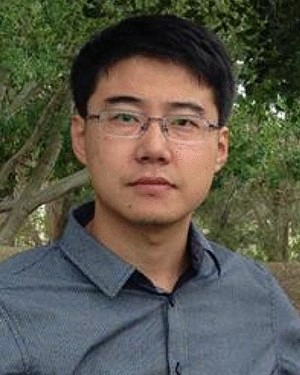}}]{Yang Yang} (Senior Member, IEEE) received the Ph.D. degree in computer science from The University of Queensland, Brisbane, QLD, Australia, in 2012. He is currently with the University of Electronic Science and Technology of China, Chengdu, China. He was a Research Fellow with the National University of Singapore, Singapore, from 2012 to 2014. His current research interests include multimedia content analysis, computer vision, and social media analytics.
\end{IEEEbiography}



\end{document}